\documentclass[10pt,twocolumn,letterpaper]{article}

\usepackage{packages/cvpr}              

\usepackage{graphicx}
\usepackage{amsmath,amsfonts,amssymb}
\usepackage{booktabs}
\usepackage{bm}
\usepackage{float} 
\usepackage{adjustbox}
\usepackage{multirow}
\usepackage{colortbl}
\usepackage{enumitem} 
\usepackage[pagebackref,breaklinks,colorlinks]{hyperref}
\usepackage{dsfont} 

\usepackage{xr}
\makeatletter
\newcommand*{\addFileDependency}[1]{
  \typeout{(#1)}
  \@addtofilelist{#1}
  \IfFileExists{#1}{}{\typeout{No file #1.}}
}
\makeatother

\newcommand*{\myexternaldocument}[1]{
    \externaldocument{#1}
    \addFileDependency{#1.tex}
    \addFileDependency{#1.aux}
}
\myexternaldocument{sections/02_method}

\usepackage[table,dvipsnames]{xcolor} 

\newcommand{\blue}[1]{\bf \color{blue!77!black}{#1}}

\newcommand{\norm}[1]{\left\lVert#1\right\rVert}

\def\cvprPaperID{9736} 
\def\confName{CVPR}
\def\confYear{2022}

\begin{document}

\title{Symmetry and Uncertainty-Aware Object SLAM \\
for 6DoF Object Pose Estimation}

\author{
Nathaniel Merrill${}^1$ \quad Yuliang Guo${}^2$ \quad Xingxing Zuo${}^3$ \quad Xinyu Huang${}^2$ \\
Stefan Leutenegger${}^3$ \quad Xi Peng${}^1$ \quad Liu Ren${}^2$  \quad Guoquan Huang${}^1$ \\
${}^1$University of Delaware \quad ${}^2$Bosch Research North America \quad ${}^3$Technical University of Munich \\
{\tt\small \{nmerrill,xipeng,ghuang\}@udel.edu \quad
\{yuliang.guo2,xinyu.huang,liu.ren\}@us.bosch.com} \\
{\tt\small \{xingxing.zuo,stefan.leutenegger\}@tum.de}
}
\maketitle

\begin{abstract}
We propose a keypoint-based object-level SLAM framework that can provide globally consistent 6DoF pose estimates for symmetric and asymmetric objects alike.
To the best of our knowledge, our system is among the first to utilize the camera pose information from SLAM to provide prior knowledge for tracking keypoints on symmetric objects -- ensuring that new measurements are consistent with the current 3D scene. 
Moreover, our semantic keypoint network is trained to predict the Gaussian covariance for the keypoints that captures the true error of the prediction, and thus is not only useful as a weight for the residuals in the system's optimization problems, but also as a means to detect harmful statistical outliers without choosing a manual threshold.
Experiments show that our method provides competitive performance to the state of the art in 6DoF object pose estimation, and at a real-time speed.
Our code, pre-trained models, and keypoint labels are available
\url{https://github.com/rpng/suo_slam}.
\end{abstract} \section{Introduction}
\label{sec:intro}

Object pose estimation in 6 degrees of freedom (DoF) plays a key role in a variety of down-stream applications (e.g., autonomous driving, robotic navigation, manipulation, and augmented reality),
and has been extensively studied in computer vision and robotics communities~\cite{Lowe1999ICCV,Rad2017ICCV,Peng2019CVPR,Park2019ICCV,Labbe2020ECCV,Fu2021IROS,Wang2021CVPR}. 
Some methods rely on RGB input~\cite{Park2019ICCV,Zakharov2019ICCV,Li2018ECCV,WangXZML0S2019CVPR,Sundermeyer2018ECCV}, while others  utilize additional depth input to improve the performance~\cite{Moreno2013CVPR,Xiang2018RSS,Li2018ECCV,WangXZML0S2019CVPR}. 
Some deal with a single view~\cite{Xiang2018RSS,Rad2017ICCV,Park2019ICCV}, 
while others utilize multiple views to enhance the results
~\cite{Collet2010ICRA,Collet2011IJRR,Moreno2013CVPR,Li2018ECCV,Labbe2020ECCV,Fu2021IROS}.
In particular, 
multi-view methods can be further categorized into offline structure from motion (SfM) -- where all the frames are given at once~\cite{Collet2010ICRA,Labbe2020ECCV} -- and the online SLAM styles, where frames are provided sequentially and real-time performance is expected~\cite{Moreno2013CVPR,Fu2021IROS}. 
This paper focuses on image-based 6DoF pose estimation for multiple objects in the context of an online monocular SLAM system.

\begin{figure}[t]
\includegraphics[width=\columnwidth,trim={0 1.5cm 0 0}]{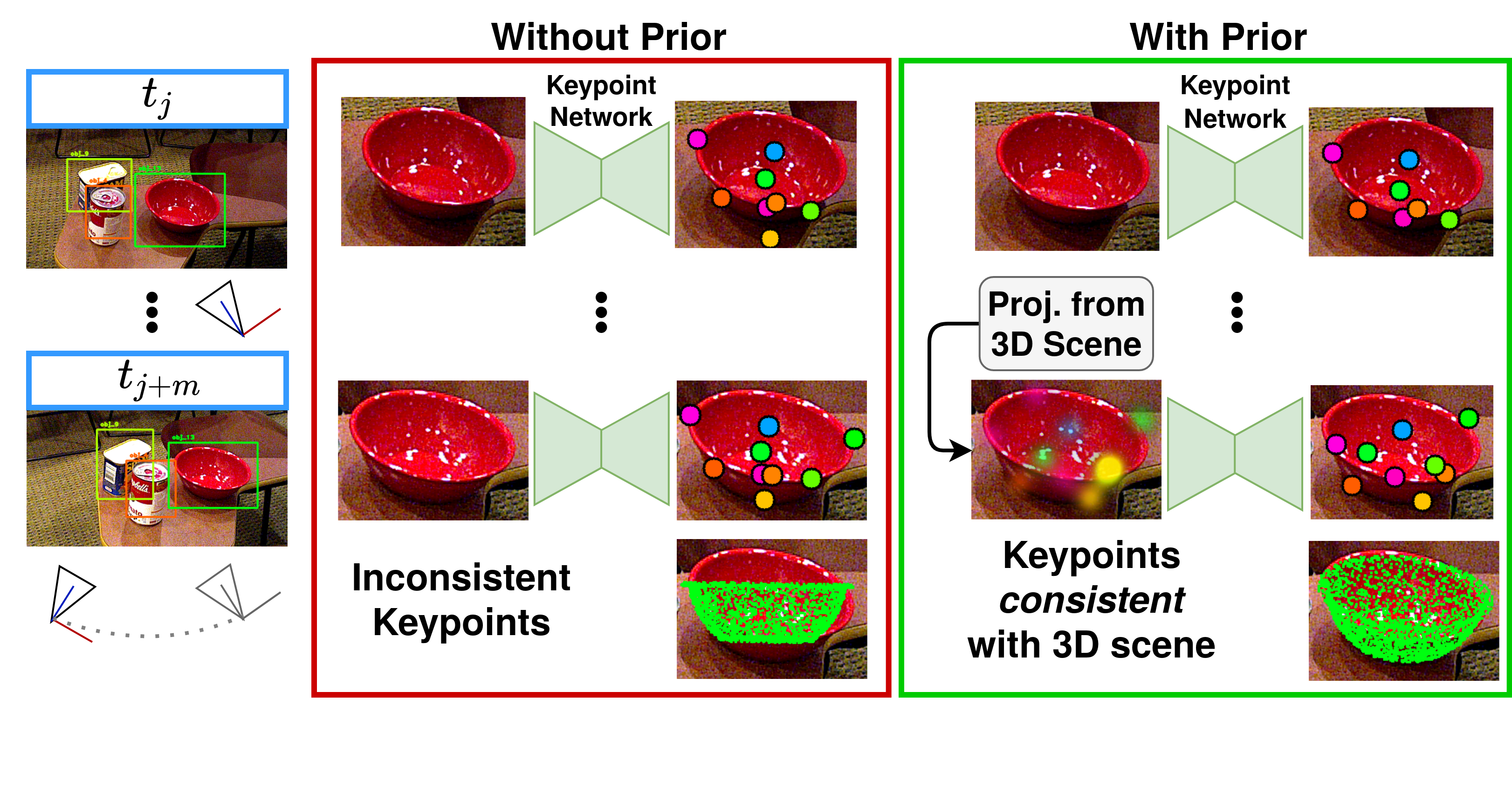}
\caption{Our proposed method leverages the detected keypoints of {\em asymmetric} objects and the 3D scene created from the SLAM system to consistently track the keypoints of {\em symmetric} objects. Given the current camera pose estimated from {\em asymmetric} objects' keypoints, the projections of the existing 3D keypoints into the current image act as informative {\em prior input} to guide the network in predicting keypoints with consistent symmetry over time.}
\label{fig:brief_overview}
\end{figure}

A typical multi-view 6DoF pose estimation method can be  decomposed into the single-view estimation stage and the multi-view enhancement stage. 
While pose estimates from multiple views can be fused for better performance~\cite{Collet2010ICRA,Labbe2020ECCV,Fu2021IROS}, handling extreme inconsistency -- e.g., those caused by rotational symmetry of objects -- is still challenging. It is also unreliable to manually tune the thresholds for outlier rejection and assign residual weights for nonlinear optimization.
To tackle these challenges, 
in this paper, we propose a symmetry and uncertainty-aware 6DoF object pose estimation method which fuses semantic keypoint measurements from all views within a SLAM framework.
The main contributions of this work are:
\begin{itemize}
    \item We design a keypoint-based object SLAM system that jointly estimates the 
          globally-consistent object and camera poses in real time -- even in the presence of
          incorrect detections and symmetric objects.
    \item We propose a method able to consistently predict and track 2D semantic keypoints for symmetric objects over time,
          which leverages the projection of existing 3D keypoints into the current image as an informative prior input to the keypoint network.
    \item We develop a method to train the keypoint network to estimate 
          the uncertainty of its predictions such that the uncertainty measure quantifies the true error of the keypoints, 
          and significantly improves object pose estimation in the object SLAM system.
\end{itemize}

The rest of this paper is organized as follows:
After briefly reviewing the related literature in Sec. \ref{sec:relwork},
we describe our method in detail in Sec. \ref{sec:method} -- including
the keypoint detector and how it is used in the entire system.
A thorough evaluation of our framework is presented in Sec. \ref{sec:exp}
before concluding in Sec. \ref{sec:conclusion}.

\section{Related Work}
\label{sec:relwork}

\paragraph{Single-view object pose estimation.}

A large number of single-view object pose estimation methods have been presented in recent years. One major trend is to utilize deep networks to predict the relative pose of an object with respect to the camera in a regress-and-refine fashion \cite{Xiang2018RSS, Li2018ECCV, Labbe2020ECCV, Zakharov2019ICCV}. Although effective, the iterative refining process is usually at a high computationally cost. Another trend is to either estimate the 2D projected locations of sparse 3D semantic points from the CAD model \cite{Rad2017ICCV, Pavlakos2017ICRA, Peng2019CVPR}, or to regress the 3D coordinates from the dense 2D pixels within object masks \cite{Park2019ICCV,Wang2021CVPR}, and then solves a perspective $n$ point (PnP) problem to estimate object poses. This type of approach is more efficient, however, not always as reliable under occlusion. In order to achieve superior robustness to occlusion and real-time efficiency simultaneously, we develop a multi-view method which integrates a sparse semantic keypoint detection in an object-level SLAM system. Instead of adapting traditional descriptors \cite{Collet2010ICRA,Collet2011IJRR}, we opt to develop a CNN-based keypoint detector in order to leverage more global context to reason about the keypoint locations and distinguish their semantics. We show that an object SLAM system can effectively utilize the sparse set of semantic keypoints to optimize the poses in a bundle adjustment (BA) optimization with outlier rejection at the keypoint level.

\paragraph{Object-level SLAM.}
Object-level SLAM typically builds upon single-view object pose estimators,
which improves the estimated poses' robustness to occlusions, missing detections, and the global consistency via multi-view optimization.
SLAM++ \cite{Moreno2013CVPR} was notably the first work along this line, but their system only worked on depth images.
There are also some works which model objects as a sparse set of 3D keypoints, and use a 2D keypoint detectors to estimate the correspondences which
are fused over time \cite{Parkhiya2018ICRA,Shan2020IROS}, however none have considered symmetric objects.
PoseRBPF \cite{Deng2019RSS} on the other hand proposed a method to track
objects over time with an autoencoder and particle filter to reason about the symmetry, however their system is only able to track one object at a time -- limiting the application.
CosyPose \cite{Labbe2020ECCV} presented a method to disambiguate pose estimates of symmetric objects from multiple views through object-level RANSAC
, but their method is an offline SfM approach and not directly comparable to ours.
Fu et. al \cite{Fu2021IROS} proposed a multi-hypothesis SLAM approach to estimate the pose of symmetric objects, which is optimized with a max-mixture model. 
In contrast, our approach only tracks one hypothesis, and is shown to have superior performance.

\paragraph{Keypoint uncertainty estimation.}
A typical global optimization that uses the predicted object keypoints as a measurement
(i.e., PnP or multi-view graph optimization), requires a proper weighting of the residuals.
Without any measure of certainty to accompany the keypoint measurements, this weight is typically set to identity or some manually-tuned value.
Some works have retrieved a weight directly from the output of the keypoint network
\cite{Pavlakos2017ICRA, Peng2019CVPR} to be used in PnP as a scalar measure of certainty \cite{Pavlakos2017ICRA} or Gaussian covariance matrix \cite{Peng2019CVPR},
while \cite{Shan2020IROS} adapted the Bayesian method of \cite{Kendall2016ArXiv}
to estimate a covariance matrix for the keypoints by sampling over a randomized batch.
Although these methods have been shown to work in practice, none have shown that the uncertainty they are predicting
actually bounds the true error of the prediction compared to the ground truth.

Besides for residual weighting, the uncertainty is especially useful for outlier rejection, since, assuming that the uncertainty is a Gaussian covariance matrix, the $\chi^2$ distribution can determine an outlier threshold more systematically compared to manual tuning.
Inspired by a plethora of recent works (unrelated to keypoint prediction) on self-uncertainty prediction of networks \cite{Bloesch2018CVPR,Klodt2018ECCV,Liu2020RAL,Yang2020CVPR,Ke2021ArXiv,Zuo2021ICRA,Matsuki2021RAL}, we design a maximum likelihood estimator (MLE) loss, which trains the network to predict keypoint locations accurately and to jointly predict the uncertainty to be tightly bound around the actual error of the prediction.
 \begin{figure*}
    \centering
    \includegraphics[width=0.85\textwidth]{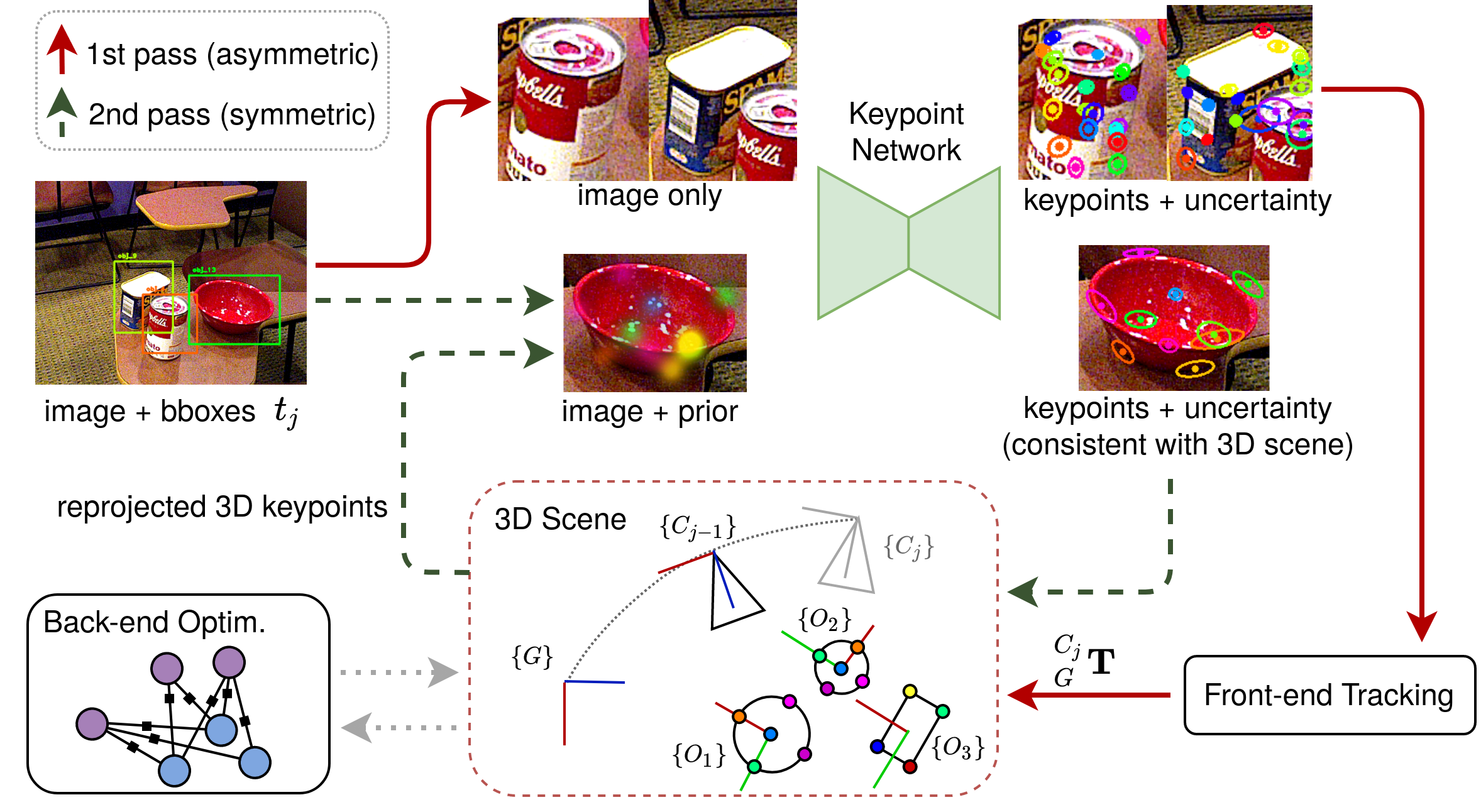}
    \caption{An overview of the proposed symmetry and uncertainty aware object SLAM pipeline.}
    \label{fig:system}
\end{figure*}

\section{The Proposed Method}
\label{sec:method}

Our multi-view 6DoF object pose estimation method is unified in an object SLAM framework, which jointly estimates object and camera poses -- while accounting for the symmetry of detected objects and utilizing the uncertainty estimations from the network to robustify the system. A depiction of the full pipeline can be seen in Fig. \ref{fig:system}. The pipeline involves two passes to deal with asymmetric and symmetric objects separately. In the first pass, the asymmetric objects are tracked from the 3D scene to estimate the camera pose. In the second pass, the estimated 3D keypoints for symmetric objects are projected into the current camera view to be used as the prior knowledge to help predict keypoints for these objects that are consistent with the 3D scene. 
The object SLAM system is primarily comprised of two modules, the front-end tracking using the keypoint network, and back-end global optimization to refine the object and camera pose estimates. 
As a result, the proposed system can operate on sequential inputs and estimate the current state in real time for the use of an operator or robot requiring object and camera poses in a feedback loop.

\subsection{Keypoint Network} \label{sec:kp_estim}

\begin{figure}[t]
    \centering
    \includegraphics[width=0.99\columnwidth,trim={0 0 4cm 0}]{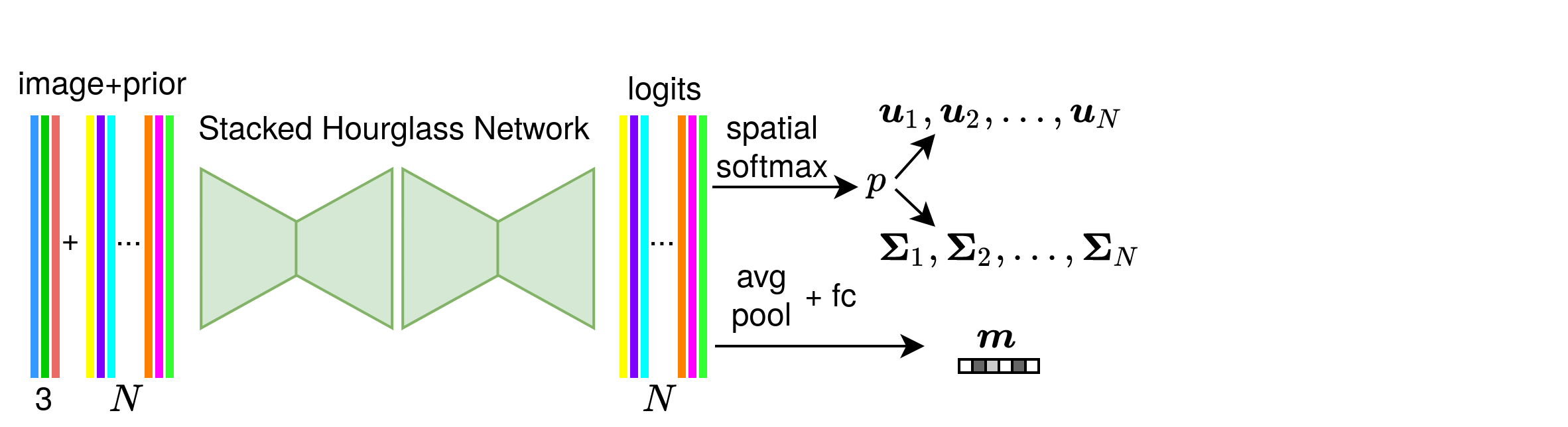}
    \caption{
        The overall architecture of our keypoint network.
        The network input is augmented to include additional $N$ channels for the prior keypoint inputs. When no prior is available, these channels are filled with zeros. 
        The network outputs an $N$-channel feature map corresponding to the raw logits, from where a spatial softmax head predicts keypoints $\bm{u}_i$ and uncertainty $\bm{\Sigma_i}$, while an average pool head predicts the keypoint mask $\bm{m}$.
    }
    \label{fig:network}
\end{figure}

We develop a keypoint network that not only predicts the 2D keypoint coordinates but also their uncertainty. In addition, to make it able to provide consistent keypoint tracks for symmetric objects, the network optionally takes prior keypoint heatmap inputs that are expected to be somewhat noisy. 
The architecture of our keypoint network can be seen in Fig. \ref{fig:network}.
The backbone architecture of our keypoint network is the stacked hourglass network \cite{Newell2016ECCV}, which has been shown to be a good choice for object pose estimation
\cite{Parkhiya2018ICRA, Pavlakos2017ICRA, Shan2020IROS}.
Similar to the original \cite{Newell2016ECCV} we choose a multi-channel keypoint parameterization due to its simplicity.
With this formulation, each channel is responsible for predicting a single keypoint, 
and we can combine all of the keypoints for the dataset into one output tensor
 -- allowing for a single network to be used for all of the objects.
 
Given the image and prior input cropped to a bounding box and resized to a static input resolution, the network predicts 
an $N \times H/d \times W/d$ tensor $p$, where $H \times W$ is the input resolution, $d$ is the downsampling ratio (4 in our experiments), and $N$ is the total number of keypoints for the dataset.
From $p$, a set of $N$ 2D keypoints $\{\bm{u}_1,\bm{u}_2,\hdots,\bm{u}_N\}$, $2 \times 2$ covariance matrices   $\{\bm{\Sigma}_1,\bm{\Sigma}_2,\hdots,\bm{\Sigma}_N\}$ are predicted.
Every channel of $p$, $p_i$, is enforced to be a 2D probability mass by utilizing a spatial softmax.
The predicted keypoint is taken as the expected value of 2D coordinates over this probability mass 
$\bm{u}_i = \sum_{u,v} p_i(u,v) [u ~ v]^\top$.
Unlike the non-differentiable argmax operation, this allows us to use the keypoint coordinate directly in the loss function -- which is important for our uncertainty estimation.

\paragraph{Keypoints with uncertainty.}
Since the keypoint $\bm{u}_i$ is the expected value of the distribution of 2D coordinates with probability mass given by the values of $p_i$, it is straightforward to estimate an uncertainty measure by the covariance of this distribution with the second moment about the mean
\begin{equation}
    \bm{\Sigma}_i = \sum_{u,v} p_i(u,v) 
    \left( [u ~ v]^\top - {\bm{u}}_i \right)
    \left( [u ~ v]^\top - {\bm{u}}_i \right)^\top.
    \label{eq:cov}
\end{equation}
However, without any particular criteria for the covariance, there is nothing to enforce that the uncertainty actually captures the true error of the prediction.
To this end, we propose to use a Gaussian maximum-likelihood estimator (MLE) loss to jointly optimize the keypoint coordinates as well as the covariance:
\begin{align} 
    L^{(i)}_\mathrm{MLE} 
    &= \left( \bm{u}^*_i - {\bm{u}}_i \right)^\top
        \bm{\Sigma}^{-1}_i 
        \left( \bm{u}^*_i - {\bm{u}}_i \right)
        + \log{|\bm{\Sigma}_i|}, 
\end{align}
where $\bm{u}^*_i$ is the ground truth keypoint coordinate.
From a high-level perspective, the first term enforces that the covariance bounds the true error of the prediction,
while the second prevents it from becoming too large.
This way, the network can predict its own uncertainty in the form of a Gaussian covariance matrix, which is trained to tightly bound the true error of the estimated keypoint.

While our network predicts a total of $N$ keypoints, only a subset of these, $\mathcal{K}(\ell) \subset \{1,2,\hdots,N\}$, are valid for a particular object $\ell$.
Furthermore, considering a single image, only a subset of keypoints $\mathcal{B} \subseteq \mathcal{K}(\ell)$ lie within the bounding box for object $\ell$ (note that occluded keypoints are still predicted).
However, during deployment, while $\mathcal{K}(\ell)$ is known from the object class and keypoint labeling, it may be impossible to know which keypoints lie within the detected bounding box.
For this reason, we add another head onto the network to predict a sigmoid vector $\bm{m} \in [0,1]^N$,
which is trained to estimate the ground-truth binary mask $\bm{m}^* \in \{0,1\}^N$, where $m^*_i = 1$ if $i \in \mathcal{B}$ and 0 otherwise (see Fig. \ref{fig:network} for the architecture).
Thus, for a single object, in a single image, the full loss becomes
\begin{align}
    L_\mathrm{tot} &= \mathrm{BCE}(\bm{m}, \bm{m^*}) +
    \frac{1}{|\mathcal{B}|} \sum_{i \in \mathcal{B}} L^{(i)}_\mathrm{MLE},
\end{align}
where $\mathrm{BCE}(.)$ is the binary cross entropy loss function.
For the rest of the paper, to simplify notation, we will denote $k \in \{1,2,\hdots,K\}$
as the indices for keypoints which pass the ground-truth mask $\bm{m}^*$ for training (i.e., the next section) or the estimated mask $\bm{m}$ (as well as the known $\mathcal{K}(\ell)$) for deployment in the SLAM system (Sec. \ref{sec:obj_slam}).

\paragraph{Keypoints for symmetric objects.}
Since we want to efficiently track the keypoints over time during deployment, it is convenient to obtain keypoint predictions that have a symmetry hypothesis that is consistent with the 3D scene.
Inspired by \cite{Moolanferoze2019ArXiv}, we opt to include $N$ extra channels as input to the keypoint network which contain a prior detection of the object's keypoints.
As shown in Fig. \ref{fig:system}, during deployment in the SLAM system, the prior keypoint detections come from projecting the 3D keypoints from the global object frame 
into the current image once the corresponding camera pose is found (i.e., the second pass).
With this paradigm, there are two main issues to address: how to create training examples
of the prior detections (since the SLAM system is not run during training), and how to detect
the initial keypoints on symmetric objects when there is not yet an object pose estimate available. 
Here we describe the training scheme used to address these issues.

To create the training prior, we simulate a noisy prior detection that the SLAM system would create by projecting the 3D keypoints from the object frame into the image plane with a perturbed ground truth object pose $\delta \mathbf{T} {}^C_O\mathbf{T}^*$ (see the supplementary Sec. \ref{sec:supp:notation} about the notation).
To further ensure that the network can learn to follow the prior detections for the symmetry
hypothesis, we utilize the set of symmetry transforms 
$\mathcal{S} = \{{}^O_{S_1}\mathbf{T}, {}^O_{S_2}\mathbf{T}, \hdots, {}^O_{S_M}\mathbf{T}\}$ that we expect to be available for each object (discretized for objects with continuous axes of symmetry).
Each ${}^O_{S_m}\mathbf{T} \in \mathcal{S}$, when applied to the object CAD model, makes the rendering look (nearly) exactly the same, and in practice, these transforms can be manually chosen fairly easily.
Thus, when constructing a training example with a prior detection, we pick a random symmetry transform and apply it to the ground-truth object pose before doing the projection.

In order for the network to learn to predict initial keypoints on symmetric objects (when no prior is available),
we only provide this simulated prior randomly for roughly half of the examples.
Without the prior detection, however, the network is left up to its own devices to reason about the absolute orientation for the object -- which is theoretically impossible for symmetric objects without special care.
As opposed to the mirroring technique and additional symmetry classifier proposed by  \cite{Rad2017ICCV}, we instead teach the network to deal with this issue with a simple criteria of choosing the keypoints that correspond
to the symmetrically-valid pose that is closest to a canonical view where the front of the object faces the camera, and the top of the object faces the top of the image.
We refer the reader to the supplementary material (Sec. \ref{sec:supp:np_choice}) for more details on this procedure.

\subsection{Object SLAM System} \label{sec:obj_slam}
Our symmetry and uncertainty-aware object SLAM system is comprised of two modules: the
front-end tracking, and the back-end global optimization.
The front end is responsible for processing the incoming frames -- running the keypoint network, estimating the current camera pose, and initializing new objects -- while the back end is responsible for refining the camera and object poses for the whole scene.
We refer the reader again to Fig. \ref{fig:system} for a visual representation of our system.

\paragraph{Front-end tracking.}
The first step of our front end is to split the bounding boxes detected in the current image into two information streams -- the first for asymmetric objects and first-time detections of symmetric ones, and the second for symmetric objects that already have 3D estimates. 
Again, we expect the symmetry information (i.e., symmetric or \textit{not}) to be included with each object class.
The first information stream sends the images, cropped at the bounding boxes, to the keypoint network without any prior to detect keypoints and uncertainty.
These keypoints are then used to estimate the pose of each asymmetric object ${}^C_O\mathbf{T}_\mathrm{pnp}$ in the current camera frame by using PnP with RANSAC.
These PnP poses are then used to coarsely estimate the current camera pose
and then initialize objects which do not yet have 3D estimates. 
See the supplementary material Sec. \ref{sec:supp:front_end} for more details on how this is done as well as more detailed behavior of the front end.

With a rough estimate of the current camera, we move onto the second information stream of the front end.
We use the coarse estimate of the camera pose to create the prior detections for the keypoints of symmetric objects by projecting the 3D keypoints for these objects into the current image, and constructing the prior keypoint heatmaps for network input.
After running the keypoint network on these symmetric objects, we store the keypoint measurements from both information streams for later use in the global optimization.

\paragraph{Back-end global optimization.}
The global optimization step runs periodically to refine the whole scene (object and camera poses) based on the measurements from each image.
Rather than reduce the problem to a pose graph (i.e., using relative pose measurements from PnP), we keep the original noise model of using the keypoint detections as measurements, which allows us to weight each residual with the covariance prediction from the network.
The global optimization problem is formulated by creating residuals that constrain
the pose  ${}^{C_j}_G\mathbf{T}$ of image $j$ and the pose ${}^{G}_{O_\ell}\mathbf{T}$ of object $\ell$ with the $k$th keypoint 
\begin{align}
    \bm{r}_{j,\ell,k} = \bm{u}_{j,\ell,k} 
    - \Pi_{j,\ell}\left( {}^{C_j}_G\mathbf{T} ~ {}^{G}_{O_\ell}\mathbf{T} ~ {}^{O_\ell}\bar{\bm{p}}_k \right),
    \label{eq:res}
\end{align}
where $\Pi_{j,\ell}$ is the perspective projection function for the bounding box of object $\ell$ in image $j$.
Thus the full problem becomes to minimize the cost over the entire scene
\begin{align}
    C 
    &= \sum_{j,\ell,k} s_{j,\ell,k} ~ \rho_H \left( 
    \bm{r}_{j,\ell,k}^\top ~ \bm{\Sigma}_{j,\ell,k}^{-1} ~ \bm{r}_{j,\ell,k}
    \right) \label{eq:global_cost} 
\end{align}
where $\bm{\Sigma}_{j,\ell,k}$ is the $2 \times 2$ covariance matrix predicted by the network for the keypoint  $\bm{u}_{j,\ell,k}$, $s_{j,\ell,k} \in \{0,1\}$ is a constant indicator 
that is 1 if the measurement was deemed an inlier before the optimization started and 0 otherwise, and  $\rho_H$ is the Huber norm which reduces the effect of outliers during the optimization steps.
Both $\rho_H$ and $s_{j,\ell,k}$ use the same outlier threshold $\tau$
, which is derived from the 2-dimensional $\chi^2$ distribution, and is always set to the 95\% confidence threshold $\tau = {5.991}$.
Thus we do not need to manually tune the outlier threshold as long as the covariance matrix $\bm{\Sigma}_{j,\ell,k}$ can properly capture the true error of keypoint $\bm{u}_{j,\ell,k}$.

\section{Experiments}
\label{sec:exp}
Our experiments are conducted on two of the most challenging object pose estimation datasets: the YCB-Video dataset \cite{Xiang2018RSS}
and the T-LESS dataset \cite{Hodan2017WACV}.
Both datasets provide ground truth poses for symmetric and asymmetric objects in cluttered environments over multiple keyframe sequences.
YCB-Video contains 21 household objects, including 4 objects with discrete symmetries and one object (the bowl) with a continuous axis of symmetry.
The T-LESS dataset contains 30 industry-relevant objects with very little texture, and most are symmetric.
Note that the symmetry information of each object is provided by \cite{Hodan2018ECCV}.

\subsection{Implementation Details} \label{sec:impl}
\paragraph{Choice of keypoints.}
While our design is agnostic to the choice of keypoint, to reduce the number of channels that the network needs to predict,
we created a set of rules to annotate keypoints manually in such a way that each keypoint can be applied to multiple object instances, and the same rules can be applied to both the YCB-Video and T-LESS dataset.
We manually label the 3D CAD models for both datasets, and project the keypoints from 3D to 2D to create the ground-truth keypoints described in Sec. \ref{sec:kp_estim}. 
We refer the reader to the supplementary material Sec. \ref{sec:supp:labeling} for more details on how we annotated the keypoints.

\paragraph{Training procedure.}
We implemented the keypoint network in PyTorch \cite{Paszke2019NeurIPS}.
For all training, we used the Adam optimizer \cite{Kingma2015ICLR} with a learning rate of $10^{-3}$.
For the YCB-Video dataset, we utilized real training data provided along with the official 80k synthetic images.
Due to the high redundancy in the real training data, we used only every 5th image.
We trained on this dataset for 60 epochs using a batch size of 24 
with randomized backgrounds for the synthetic dataset as well as randomized bounding boxes, color, and image warping.
For the T-LESS dataset, there are only real training images of single objects on a dark background
, so for the synthetic data
we opted to use the physics-based \texttt{pbr} rendered data provided by \cite{Hodan2020ECCVW}.
For both the \texttt{real} and \texttt{pbr} splits we augment the examples with randomized backgrounds, bounding boxes, color, and warping, as well as randomly pasted objects for the real data only -- since it only contains images of isolated objects.
We trained the TLESS model for 89 epochs with a batch size of 8, which was smaller
than that for YCB-Video due to the higher image resolution of the \texttt{pbr} data.

\paragraph{SLAM system.}
Our SLAM system is implemented in Python.
The GPU is only used for network inference while all other operations are performed on the CPU.
All optimizations are implemented using Python wrappers for the g2o library \cite{Kuemmerle2011ICRA}\footnote{\url{https://github.com/uoip/g2opy}}, besides PnP,
which is done using the Lambda Twist solver \cite{Persson2018ECCV} with RANSAC\footnote{\url{https://github.com/midjji/pnp}}.
Our front-end tracking works on every incoming frame, while the back-end runs every 10th frame.
Note that the testing sequences for both datasets are already provided as keyframes, so no keyframing procedure is needed.
While for actual deployment it is ideal to run the back-end graph optimization on a separate worker thread, this would make reproducing the exact results impossible due to randomness in the operating system's allocation of resources between the two threads.
In order to make the results reproducible, we simply execute both the front-end and back-end on the main thread for evaluation.
Our front-end tracking can typically run at 11Hz on our desktop with a GTX 1080Ti graphics card, and the back-end can run at an average speed of 2Hz.

\subsection{YCB-Video Dataset} \label{sec:ycbv}

\begin{table}
\centering
\caption{Results on the YCB-Video dataset. Data means what synthetic data was used in addition to the real data, and U.M. (unified model) is checked if only one model was trained for all objects instead of one model trained for each object separately. Bold is best, underlined is second best.
}
\begin{tabular}{l|cccc}
\toprule
Method & Data & U.M. & ADD-S & ADD(-S) \\
\hline
PoseCNN \cite{Xiang2018RSS} & \texttt{syn} & \checkmark & 75.3 & 61.3 \\
DeepIM \cite{Li2018ECCV} & \texttt{syn} & \checkmark & 88.1 & 81.9 \\
PoseRBPF \cite{Deng2019RSS} & \texttt{syn} &  & 76.3 & 64.4 \\
MHPE \cite{Fu2021IROS} & \texttt{syn} & \checkmark & 82.9 & 69.7 \\
CosyPose \cite{Labbe2020ECCV} & \texttt{pbr} & \checkmark & 89.8 & \underline{84.5} \\
GDR-Net \cite{Wang2021CVPR} & \texttt{pbr} & \checkmark & 89.1 & 80.2 \\
GDR-Net \cite{Wang2021CVPR} & \texttt{pbr} &  & {\bf 91.6} & 84.4 \\
\hline
Ours & \texttt{syn} & \checkmark & \underline{90.3} & {\bf 84.7} \\ 
~ no prior det & \texttt{syn} & \checkmark & 88.7 & 83.3 \\
~ manual cov & \texttt{syn} & \checkmark & 59.1 & 46.1 \\
~ no MLE loss & \texttt{syn} & \checkmark & 47.0 & 35.2 \\
~ single view & \texttt{syn} & \checkmark & 65.7 & 56.9 \\
\bottomrule
\end{tabular}
\label{table:ycbv_main}
\end{table}

\begin{figure}[t]
    \centering
    \includegraphics[width=0.9\columnwidth]{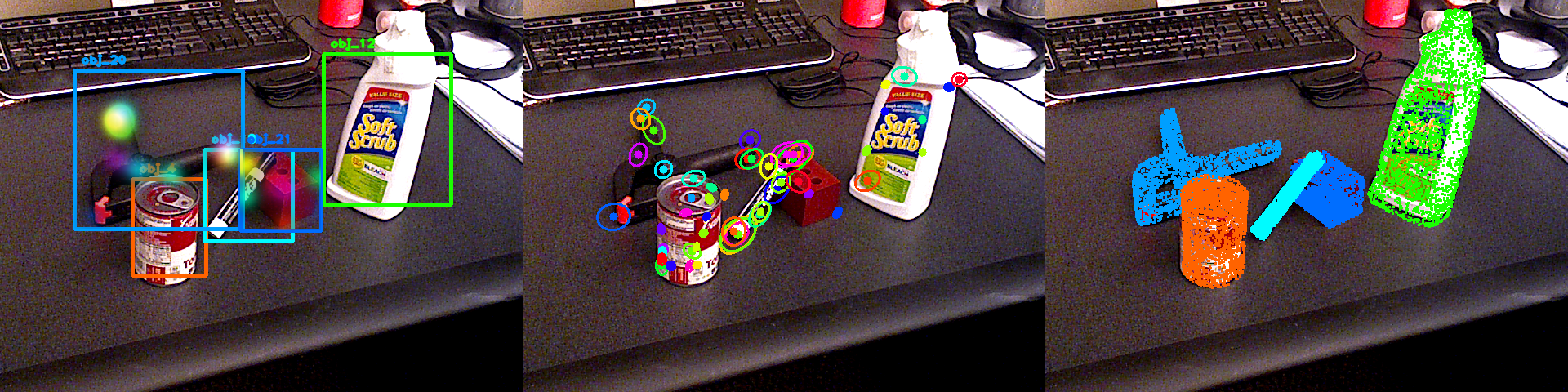}
    \includegraphics[width=0.9\columnwidth]{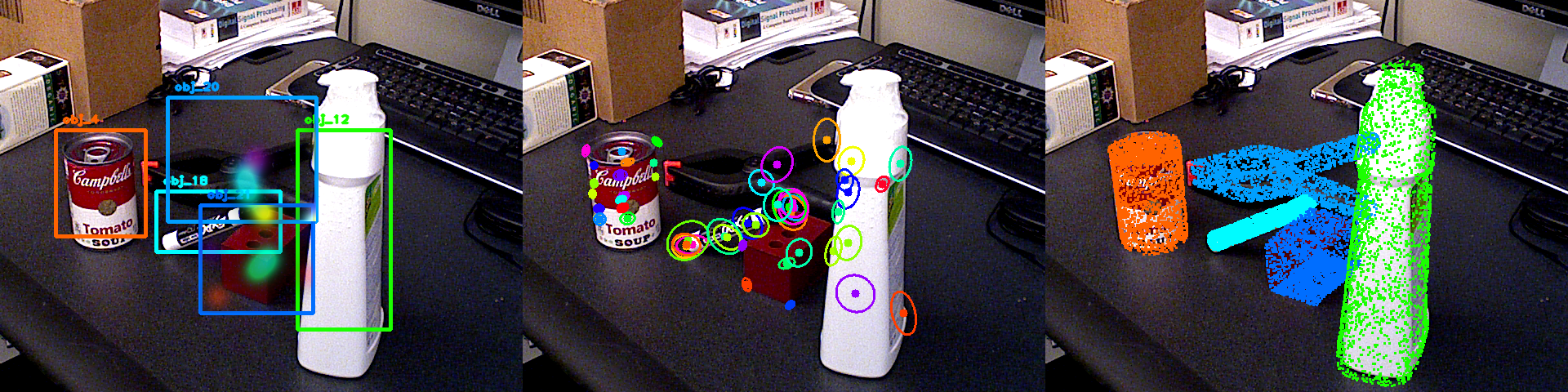}
    \includegraphics[width=0.9\columnwidth]{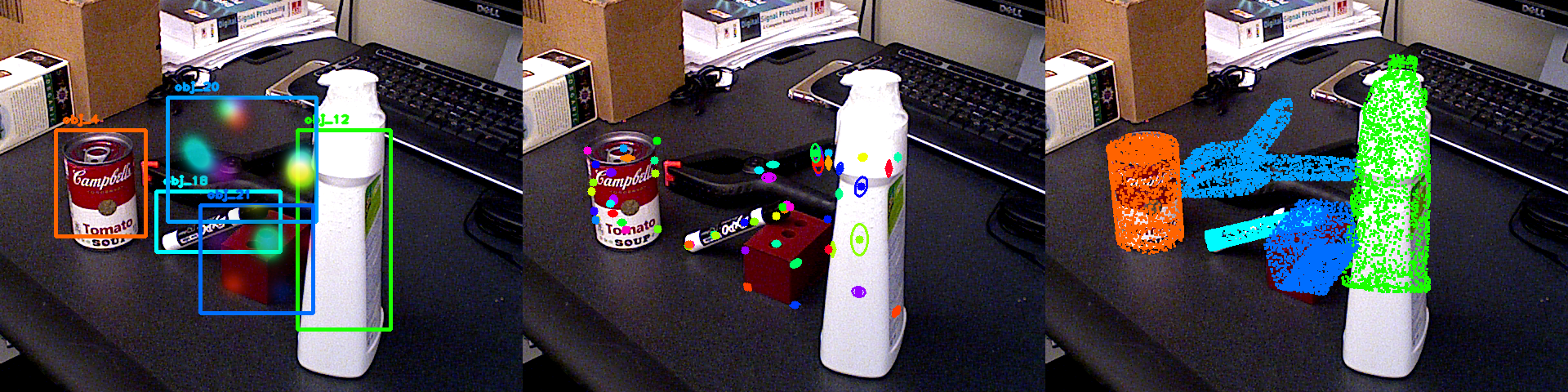}
    \caption{Qualitative results on YCB-Video. From left to right columns we show the detected object boxes with prior input to the keypoint network, the predicted keypoints with uncertainty ellipses, and the 3D model projection on the image based on predicted 6Dof object poses and camera pose. 
    {\bf Top:} the uncertainty ellipses tend to be smaller for visible keypoints on textured surfaces or corner points, while appearing larger for occluded keypoints and keypoints on smooth surfaces (like the clamp).
    {\bf Center:} our system is able to consistently track the keypoints throughout the scene despite the presence of symmetric objects.
    {\bf Bottom:} the network trained with a fixed-variance loss predicts uncertainty ellipses that are visibly too small -- leading to unreliable outlier rejection and object poses.
    }
    \label{fig:ycbv_qual}
\end{figure}

For the YCB-Video dataset, we compare to the single-view methods \cite{Xiang2018RSS,Li2018ECCV,Labbe2020ECCV,Wang2021CVPR} and SLAM methods \cite{Deng2019RSS,Fu2021IROS}.
Note that we do not include the multi-view results of CosyPose \cite{Labbe2020ECCV} since it is an offline SfM method that is not comparable to real-time SLAM methods.
Following \cite{Xiang2018RSS,Li2018ECCV,Labbe2020ECCV,Wang2021CVPR,Deng2019RSS,Fu2021IROS}, we report the area under curve (AUC) of the ADD-S and ADD(-S) by varying the accuracy threshold from 0 to 10cm, which is calculated for each object separately and then averaged.
To fairly compare the methods, we used the same bounding boxes as PoseCNN.
In practice, the bounding boxes can come from any real-time bounding box detector.
The benchmark results as well as several ablation studies are reported in Table \ref{table:ycbv_main} with our method labeled as ``Ours''.
Methods in Table \ref{table:ycbv_main} are marked as using standard synthetic data (\texttt{syn}) with randomly-placed objects or physics-based (\texttt{pbr}) training data in addition to the real data.
Note that while the \texttt{pbr} data is generally considered
superior to the randomly-placed objects \cite{Hodan2020ECCVW},
it is not a part of the official YCB-Video dataset training splits.
Regardless, our method beats all of the state-of-the-art single view and SLAM methods in terms of the AUC of ADD(-S) metric -- even those utilizing the \texttt{pbr} data while only utilizing one network
for all objects.
The AUC of ADD(-S) is the most important metric here, since it 
takes into account the actual object symmetries rather than just shape matching like the ADD-S does.
This shows that our system can provide highly accurate globally-consistent poses for symmetric objects, while still maintaining high accuracy on the texture-asymmetric objects.
Qualitative results can be seen in Fig. \ref{fig:ycbv_qual}.
More detailed results of each object category can be found in the supplementary material Sec. \ref{sec:supp:ext}.

\paragraph{Effect of prior detection.}
The first ablation study is to run our same system without the prior detection.
The results drop slightly, but this is expected on this dataset where only 5 out of 21 objects are considered symmetric, and only the bowl displays a continuous rotational symmetry.
In the next section, we will see that the prior detection actually makes a much bigger difference on the T-LESS dataset, where most of the objects are symmetric, and the camera rotates many times completely around the scene -- whereas the camera motion in YCB-Video is much simpler.

\paragraph{Manual covariance weight.}
For the next ablation in Table \ref{table:ycbv_main}, ``manual cov'', we manually tune a weight to replace the covariance in the SLAM system's residuals and outlier rejection mechanism. 
Here, we found that the weight corresponding to $2\times$ the average predicted standard deviation of the network (which was about 2.5 pixels) achieved the best scores. 
As observed, the results dropped significantly compared to using a network predicted covariance.

\paragraph{Effect of MLE loss.}
For the ablation labeled ``no MLE loss'', we trained a network with the same procedure, but replaced the MLE loss with a fixed-variance loss with variance regulation 
similar to that used by the popular human pose estimation \cite{Nibali2018ArXiv}.
As observed, when placed in the SLAM system, the results are significantly lower than that with our network trained with the MLE loss.
The qualitative results of this experiment are also in Fig. \ref{fig:ycbv_qual}.

\begin{figure}[t]
    \centering
    \includegraphics[width=0.48\columnwidth]{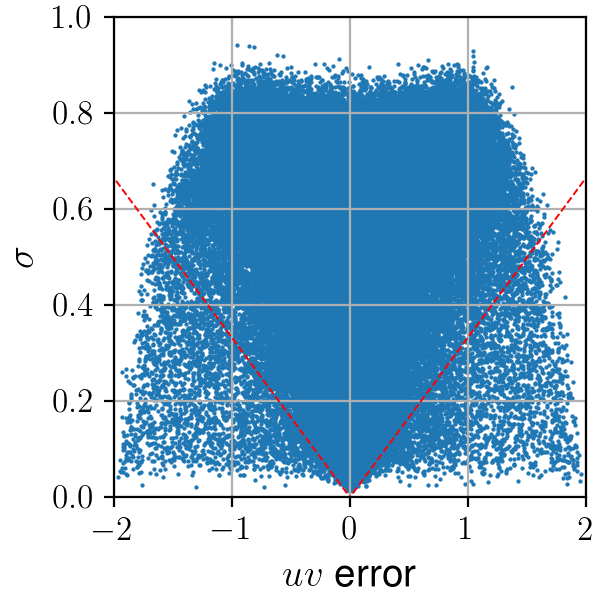}
    \includegraphics[width=0.48\columnwidth]{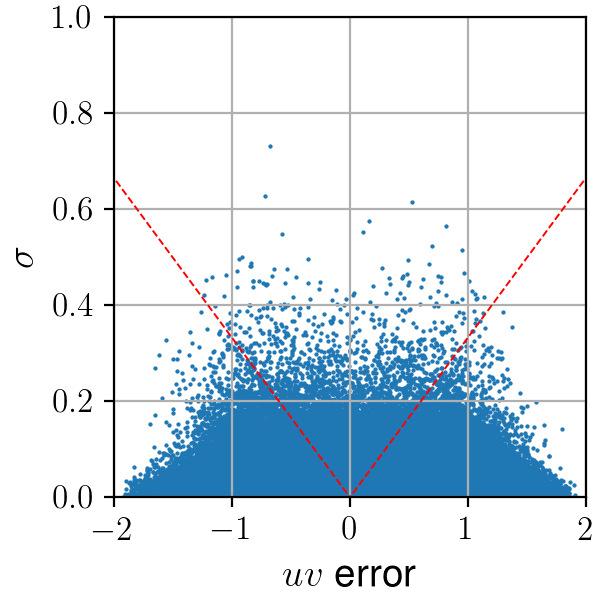}
    \caption{
    The plot of error of the predicted keypoints against the standard deviation predicted by the network over a separate set of rendered YCB-Video objects. The 3$\sigma$ bounds are shown as the cone drawn as the red dotted lines. \textbf{Left:} The result of our network trained with the MLE loss. \textbf{Right:} The result of the same network trained with a typical fixed variance loss instead, which have far fewer points within the 3$\sigma$ cone.  
    }
    \label{fig:cov_consist}
\end{figure}

Beyond the accuracy of the SLAM system with this change, we have also tested the accuracy of the predicted covariance itself.
To do so, we ran both of the networks (with and without MLE loss) on a separate set of simulated YCB-Video objects (the \texttt{pbr} data which was not used in training), which has perfect ground truth for the keypoints.
Here, we ran the networks with the ground truth bounding boxes and no prior detection.
To evaluate the accuracy of the predicted covariance, we plotted the keypoint error 
against the predicted standard deviation of the network.
Ideally, the error will always lie above the cone $e_r < 3 \sigma$ if $e_r$ is the scalar $x$ or $y$ component of the error residual of the keypoint prediction. 
The results of this experiment can be viewed in Fig. \ref{fig:cov_consist}.
As observed, the network trained with the MLE loss has much more of the errors within the 3$\sigma$ cone.
In fact, 91.0\% of the data points on the left in Fig. \ref{fig:cov_consist} pass a 99\% confidence $\chi^2$ test while only 7.1\% pass from the points on the right. This shows that the predicted uncertainty describes the actual error distribution well (besides some expected outliers due to heavy occlusion and symmetry), and including the MLE loss is crucial to achieve this.

\paragraph{Comparing to single view.}
For the final ablation in Table \ref{table:ycbv_main}, we ran just our single view network and compared the accuracy.
Specifically, for each view we just ran PnP and refined it using the same procedure as Eq. \ref{eq:global_cost}, but with only one fixed camera pose per optimization.
Clearly the full SLAM system is more accurate.
It is interesting to note that the results for single view are actually more accurate than the SLAM results using the manual covariance or the fixed-variance network.
This is most likely due to the fact that incorrect covariance in our SLAM system can cause the outlier rejection mechanism to be unreliable, and outliers can then pull the object pose in an incorrect direction and hurt the accuracy for {\em all views} despite the fact that most of the keypoints are correct.

\paragraph{Accuracy of camera poses}
The effect of initializing the camera poses with the poses provided by the dataset was minor in this experiment. 
Using the given camera poses the system achieved a 90.5 AUC of ADD-S score, while the 
system with the estimated camera poses scored the 90.3 shown in Table \ref{table:ycbv_main}.
This shows that the estimated camera poses are very accurate on this dataset.

\subsection{T-LESS Dataset} \label{sec:tless}

\begin{table}
\centering
\caption{Benchmark Results on the T-LESS dataset.}
\begin{tabular}{l|ccc}
\toprule
Method & Data & U.M. & $e_\mathrm{vsd} < 0.3$ \\
\hline
Implicit \cite{Sundermeyer2018ECCV} & \texttt{syn} &  & 26.8 \\
Pix2Pose \cite{Park2019ICCV} & \texttt{syn} &  &  29.5 \\
PoseRBPF \cite{Deng2019RSS} & \texttt{syn} &  & 41.7 \\
CosyPose \cite{Labbe2020ECCV} & \texttt{pbr} & \checkmark  & {\bf 63.8} \\
\hline
Ours & \texttt{pbr} & \checkmark & \underline{63.7} \\ 
~ \texttt{real} only & N/A & \checkmark & 45.9 \\ 
~ no prior det & \texttt{pbr} & \checkmark & 16.2 \\
~ manual cov & \texttt{pbr} & \checkmark & 13.8 \\
~ single view & \texttt{pbr} & \checkmark & 48.1 \\
\bottomrule
\end{tabular}
\label{table:tless_main}
\end{table}

\begin{figure}[t]
    \centering
    \includegraphics[width=0.9\columnwidth]{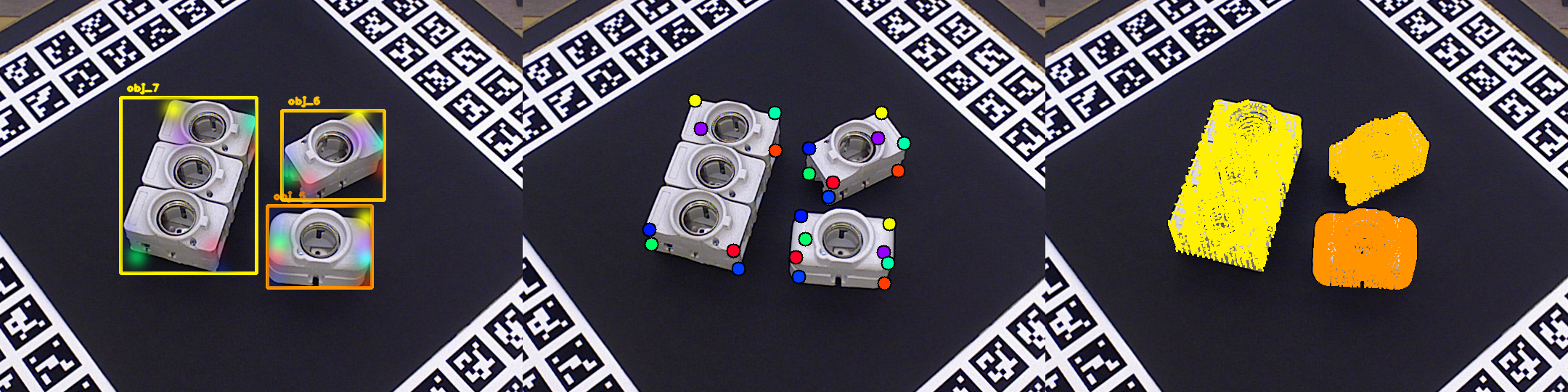}
    \includegraphics[width=0.9\columnwidth]{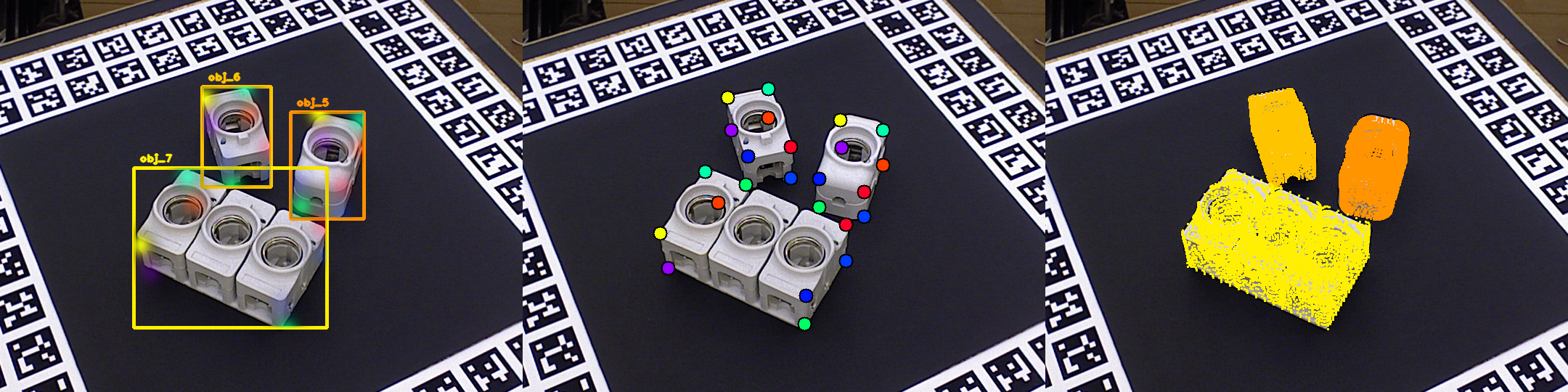}
    \includegraphics[width=0.9\columnwidth]{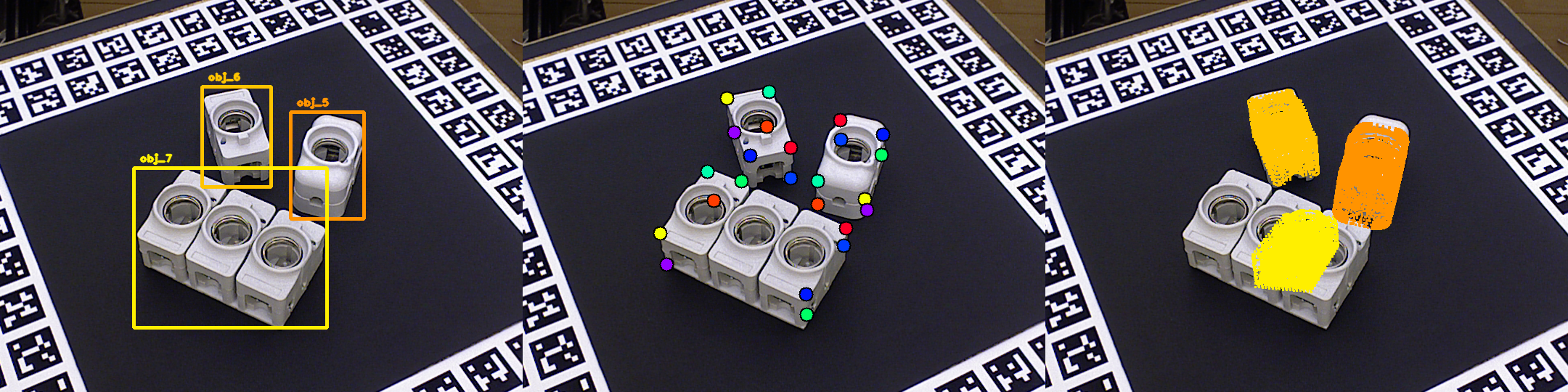}
    \caption{Qualitative results on T-LESS. {\bf Top:} Under misalignment between the prior detection and objects (left column), the network still predicts keypoints accurately (center column) which just uses the prior as a general guide for the symmetry.
    {\bf Center:} the system displays robustness to missing and bad bounding boxes here.
    {\bf Bottom:} the same system, but without the prior detection, fails to track keypoints corresponding to the same 3D locations currently locating at the back side of the symmetric objects, hence causes the estimated object poses to fly away.
    Note that the predicted covariance was used in all of these images, but left out of the visualization for clarity.
    Best viewed in color.
    }
    \label{fig:tless_qual}
\end{figure}

For the T-LESS dataset, we compare to two single-view baselines \cite{Sundermeyer2018ECCV, Park2019ICCV} as well as, again, PoseRBPF \cite{Deng2019RSS} and CosyPose \cite{Labbe2020ECCV}.
To fairly compare to the other methods, we use the same RetinaNet bounding boxes as \cite{Park2019ICCV},
taking the top scoring bounding box for each object.
We use the standard visual surface discrepancy (vsd) recall metric,
$e_\mathrm{vsd} < 0.3$ \cite{Hodan2018ECCV}, that the other methods reported.
Since the T-LESS dataset has multiple scenes that have only symmetric objects, and 
our system requires asymmetric objects to estimate a camera pose, we initialize our camera poses with the poses provided by the dataset.
While this is a potential drawback to our system, typical deployment scenarios will contain symmetric objects or allow for retrieving external odometry from another source, such as an additional IMU sensor or traditional feature-based SLAM.

The benchmark results and ablation studies are reported in Table \ref{table:tless_main}, where our system is shown to achieve a 63.7 recall score -- second best to the 63.8 of CosyPose.
However, it is interesting to note that CosyPose is an iterative refinement method that utilizes initial object poses rendered at 1m from the camera, which is close to the distance of all the objects, while our method makes no such assumption.
Qualitative results can be also seen in Fig. \ref{fig:tless_qual}.

\paragraph{Effect of training data.}
To test the sensitivity to the training data, 
we train it on only the small \texttt{real} training split, 
which contains 1,231 images of each object on a dark background.
From Table \ref{table:tless_main} we observe that, even with this small amount of data,
we still beat all of the state-of-the-art methods besides CosyPose -- all of which used large amounts of 
synthetic data on top of the real data.
This shows the ability of our method to work with a limited amount of data which does not even cover all 
orientations of the objects.

\paragraph{Effect of prior detection.}
On the T-LESS dataset, where most of the objects are symmetric in some way,
the 63.7 recall score drops to 16.2 in Table \ref{table:tless_main} when the prior detection is removed.
This shows that the prior detection is crucial for tackling these challenging T-LESS objects when the camera is orbiting around their axes of symmetry multiple times.
Without the prior detection, the SLAM system's outlier rejection simply rejects most of the keypoint measurements on the symmetric objects, as they do not correspond to the same 3D location. 
Fig. \ref{fig:tless_qual} also includes some qualitative results of this experiment.

\paragraph{Manual covariance weight.}
Here again we set the covariance in the SLAM system's residuals to a manually-tuned weight.
The result in this case drop to a 13.8 recall, which further substantiates
the usefulness of our covariance estimate in the SLAM system.
Furthermore, we found that the optimal weight for this dataset was much larger than
that for YCB-Video, which is not surprising, but shows that removing the need
to manually tune weights by using the predicted covariance is a useful property of our system.

\paragraph{Comparing to single view.}
In this case, the single view result in Table \ref{table:tless_main} outperformed that from the SLAM system when it either used a manual covariance weight or no prior detections.
Since the single-view results use no prior detection, this shows that the keypoints considered independently for each view are reasonable, while the prior detection is crucial for tracking them across time.
 \section{Conclusions and Future Work} \label{sec:conclusion}
In this work, we have designed a keypoint-based object-level SLAM system that provides globally consistent 6DoF pose estimates for objects with or without symmetry.
Our method can track semantic keypoints on symmetric objects consistently with the aid of the proposed prior detection, and the uncertainty that our network predicts has been shown to capture the true error of the predicted keypoints as well as greatly improve the  object pose accuracy.
In the future, we would like to adapt our system to larger environments and generalize to class-level keypoint prediction with unseen instances.

\paragraph{Acknowledgement.}
We would like to thank the reviewers for their constructive feedback.
This work was partially supported by the University of Delaware
College of Engineering, the NSF (IIS-1924897), the ARL
(W911NF-19-2-0226, W911NF-20-2-0098), Bosch Research North America, and the Technical University of Munich. 

{\small
\bibliographystyle{packages/ieee_fullname}
\bibliography{library}
}

\cleardoublepage
\appendix
\section*{Supplementary Material} \label{sec:supp}
\section{Rigid Body Transform Notation} \label{sec:supp:notation}
Throughout the paper, we have regularly included rigid body transforms in many equations.
Here, we briefly explain the notation.
A 6DoF rigid body transform ${}_A^B\mathbf{T} \in SE(3)$ will transform a point defined in the reference frame $\{A\}$ into the reference frame $\{B\}$.
We write this in two possible ways.
For the first, and most common, we separate ${}_A^B\mathbf{T}$ into its rotational and positional components,  ${}_A^B\mathbf{R} \in SO(3)$ and  ${}^B \bm{p}_A \in \mathbb{R}^3$ respectively.
In this form we write ${}^B\bm{p}_k = {}_A^B\mathbf{R} ~ {}^A\bm{p}_k + {}^B \bm{p}_A$ to transform the 3D point ${}^A\bm{p}_k$ from the $\{A\}$ frame into the $\{B\}$ frame.

In the other form, which shows up in Eq. \ref{eq:res}, we leave the transform in its full $4 \times 4$ $SE(3)$ form,
and use the homogeneous form of translation vectors ${}^A\bar{\bm{p}}_k = [{}^A\bm{p}_k^\top ~ 1]^\top$.
In this way, we write ${}^B\bar{\bm{p}}_k = {}_A^B\mathbf{T} ~ {}^A\bar{\bm{p}}_k$.
This form specifically allows us to chain together multiple transformations with simplified notation,
for example: ${}^B\bar{\bm{p}}_k = {}_{A_2}^B\mathbf{T} ~ {}_{A_1}^{A_2}\mathbf{T} ~ {}_A^{A_1}\mathbf{T} ~ {}^A\bar{\bm{p}}_k$.

\section{Choice of Symmetry Without Prior}  \label{sec:supp:np_choice}
\begin{figure}
    \centering
    \includegraphics[width=\columnwidth]{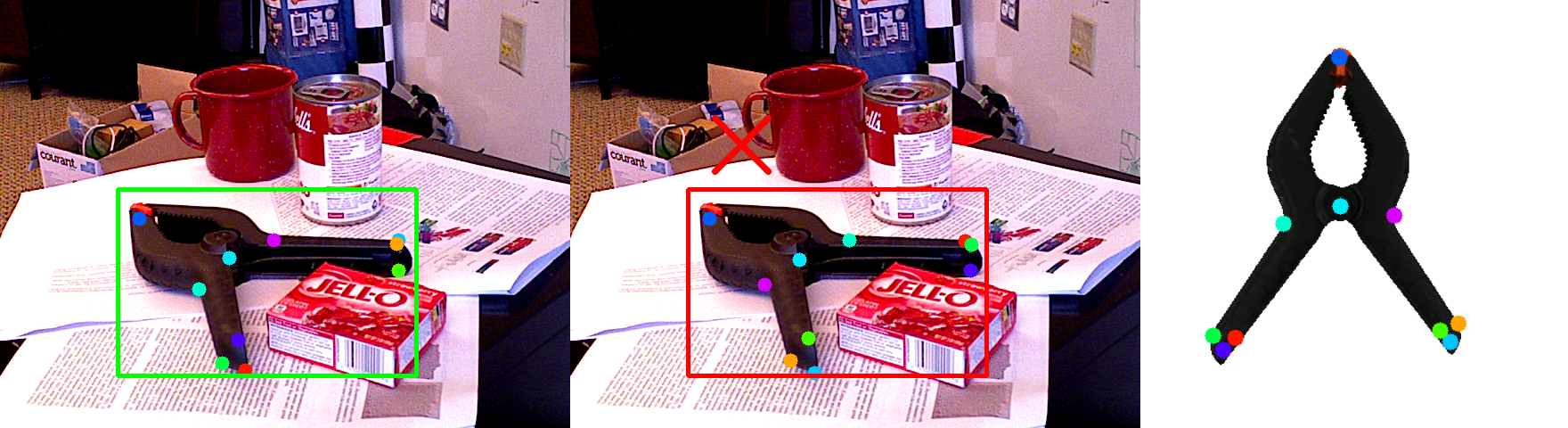}
    \includegraphics[width=\columnwidth]{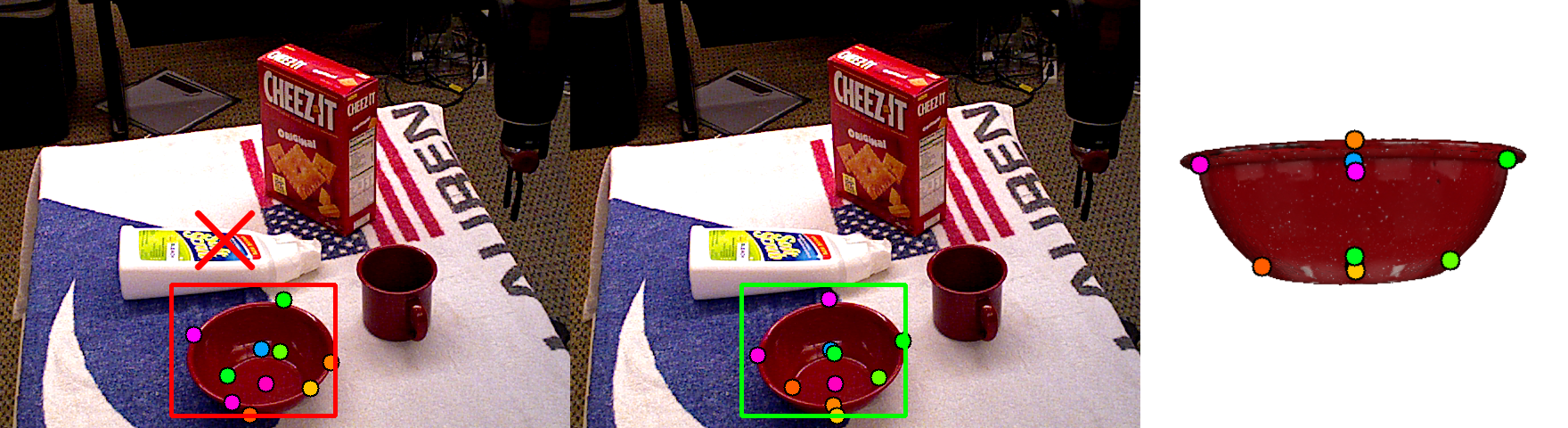}
    \caption{Examples of how we pick the symmetry to use for the keypoints during training when a prior detection is not given to the network. {\bf Top:} the two possible symmetries for the clamp are shown on the left, and the keypoints in the canonical view are shown on the right. The first symmetry is chosen since the points are closer to the points in the canonical view. 
    {\bf Bottom:} the bowl has a continuous axis of symmetry about its vertical axis which are discretized into 64 symmetry transforms. For brevity we only show two -- a random symmetry that is not chosen for the label (left) and one that is chosen (center) since it matches the canonical view (right) the best in terms of orientation.
    Best viewed in color.}
    \label{fig:sym_training}
\end{figure}

As mentioned in Section \ref{sec:kp_estim}, as opposed to the mirroring technique and additional symmetry classifier proposed by  \cite{Rad2017ICCV}, we need to teach the network to predict the initial keypoints of symmetric objects, before the prior is available.
We opt to utilize the set of symmetry transforms to solve this issue in a more concise manner with a simple intuition: when the prior detection is not available for a symmetric object, we can simply instruct the network to choose the orientation which brings the object pose closest to a canonical pose where the front of the object faces the camera, and the top of the object faces the top of the image.
This intuition is learned by the network during training by choosing the symmetry for keypoint labels that brings the 3D keypoints closest (in orientation) to those transformed into the canonical view $\{O_c\}$ in the camera frame:
\begin{gather}
    {}^O_{S}\mathbf{T} = \underset{{}^O_{S_m}\mathbf{T} \in \mathcal{S}}{\mathrm{argmin}}
    \frac{1}{K} 
    \sum_{k=1}^K \norm{
        {}^C\tilde{\bm{p}}_k - {}^C\tilde{\bm{p}}^{c}_k
    }_2 \label{eq:sym_gt} \\
    {}^C{\bm{p}}_k = {}^C_{O}\mathbf{R} \left({}^O_{S_m}\mathbf{R} {}^O\bm{p}_k + {}^O\bm{p}_{S_m} \right)
    ~~~~~~~~~~ {}^C{\bm{p}}^{c}_k = {}^C_{O_c}\mathbf{R} {}^O\bm{p}_k \notag
\end{gather}
where $\tilde{\bm{p}}_k = \bm{p}_k - \frac{1}{K}\sum_{k=1}^K \bm{p}_k$ denotes the $k$th point of a mean-subtracted point cloud.
We provide some visual examples of the effect of Eq. \ref{eq:sym_gt}, which can be seen in Fig. \ref{fig:sym_training}.
Remember that Eq. \ref{eq:sym_gt} is used to pick the symmetry transform to apply to the ground truth keypoints
during training when the simulated prior detection is not given to the network -- otherwise a random symmetry 
transform is applied to the prior and ground truth keypoints together so that the network can learn to follow the
prior for the symmetry.
The main effect of Eq. \ref{eq:sym_gt} is that it will choose the symmetry to apply to the ground truth keypoints that best matches the canonical view in terms of orientation, which essentially tells the network to always pick the symmetry that brings the front of the object closest to the camera and the top of the object closest to the negative $y$-axis of the camera frame (i.e., the top of the images) if no prior is given.

Of course there is still the issue of detecting keypoints near the inflection point of a symmetry \cite{Rad2017ICCV}.
While we could utilize the mirroring technique of \cite{Rad2017ICCV} to avoid this issue, we only need to detect keypoints once without the prior detection in practice (i.e., the first detection), and the mirroring technique of \cite{Rad2017ICCV} requires an additional classifier during test time -- which complicates the pipeline and adds additional computation.
If the object is at an inflection point for the symmetry, and it is difficult to decide which symmetry to use, in our full SLAM system we can typically just reject bad measurements until the camera moves to a better viewpoint on the object in order for the network to more confidently choose the initial symmetry based on its training with Eq. \ref{eq:sym_gt}.

\begin{figure*}
    \centering
    \includegraphics[width=0.9\textwidth]{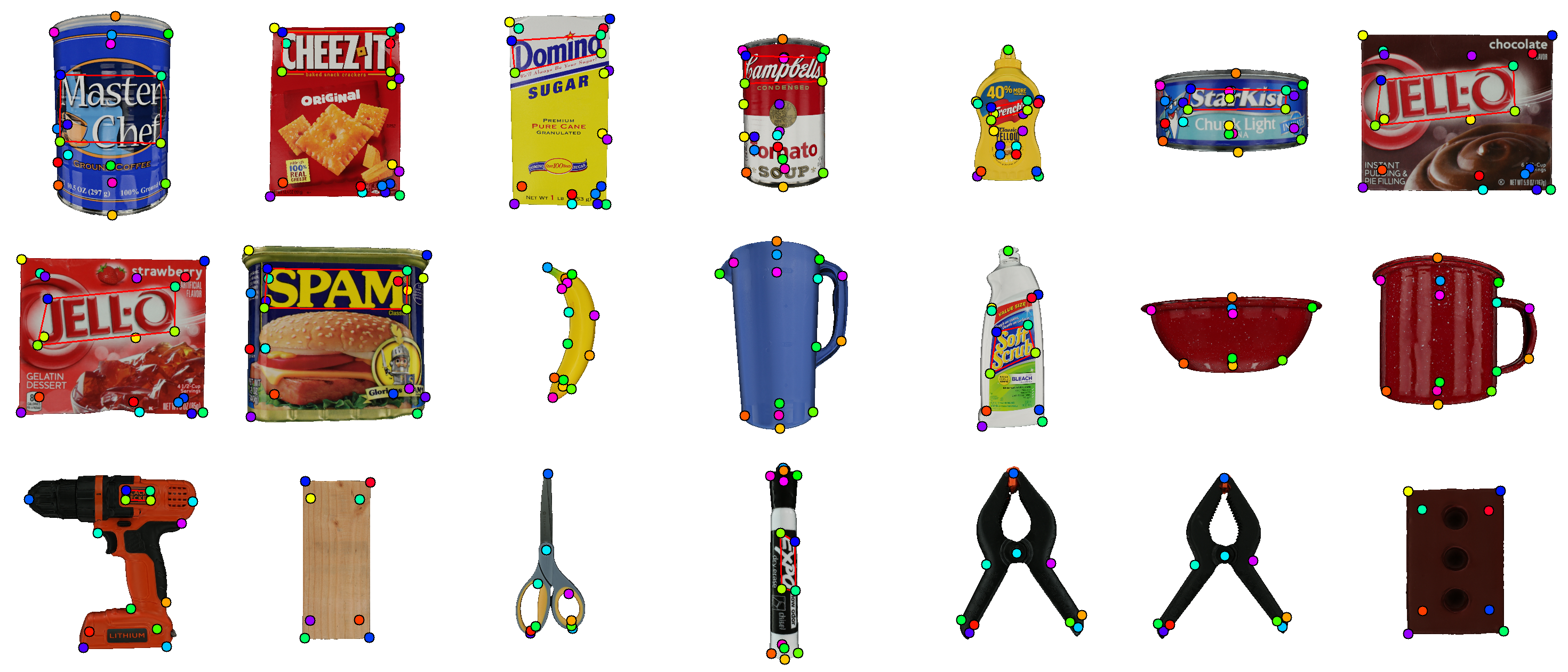}
    \caption{Our keypoint labels for the YCB-Video dataset. We labeled identifiable features based on the shape class of the objects (box-like, cylinder-like, and hand tool) which are common within different instances of the same shape class (such as box corners, cylinder top/bottom center, etc), and then instance-specific keypoints of other identifiable features such as brand names, bar codes, etc.}
    \label{fig:ycbv_labels}
\end{figure*}

\begin{figure*}
    \centering
    \includegraphics[width=0.9\textwidth]{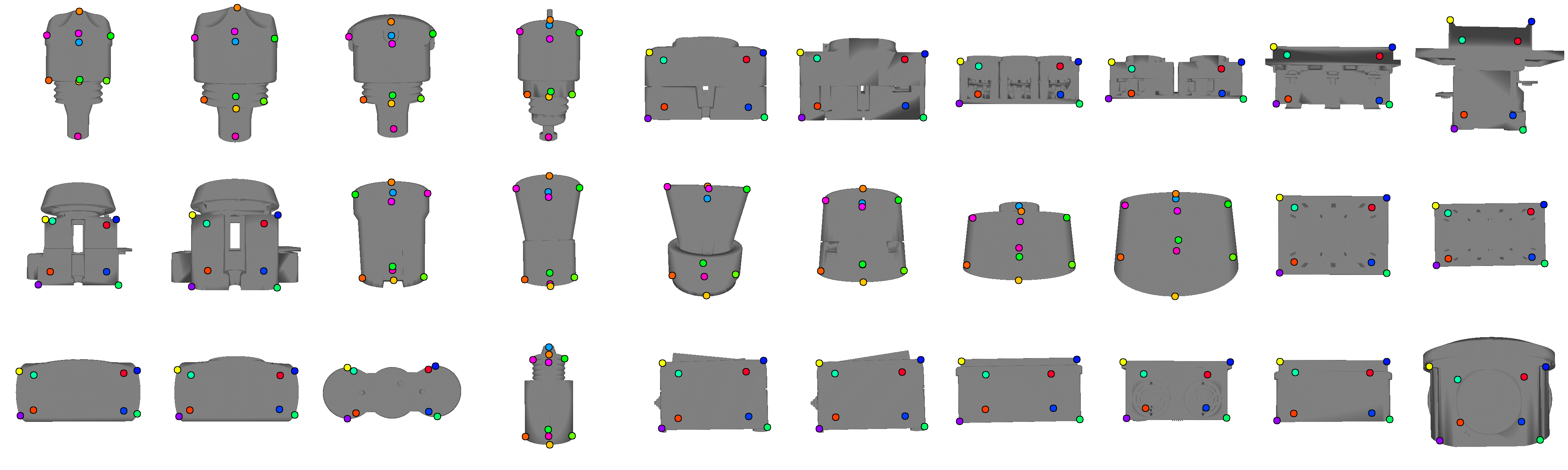}
    \caption{Our keypoint labels for the T-LESS dataset. Here, only shape class-specific keypoints were used due to the lack of texture on each object. }
    \label{fig:tless_labels}
\end{figure*}

\section{Front-End Tracking Details} \label{sec:supp:front_end}

Besides the first image, whose camera frame becomes the global reference frame $\{G\}$, we need to estimate the camera pose ${}^C_G\mathbf{T}$ with the set of object PnP poses and the current estimates of the objects in the global frame.
For each asymmetric object that is both detected in the current frame with a successful PnP pose ${}^C_O\mathbf{T}_\mathrm{pnp}$ and has an estimated global pose ${}^G_O\mathbf{T}$, we can create a hypothesis about the current camera's pose as
${}^C_G\mathbf{T}_\mathrm{hyp} = {}^C_O\mathbf{T}_{pnp} ~ {}^G_O\mathbf{T}^{-1}$ 
and then project the 3D keypoints from all objects that have both a global 3D estimate and detection in the current image into the current image plane with this camera pose, and count inliers with a $\chi^2$ test using the detected keypoints and uncertainty.
We take the camera pose hypothesis with the most inliers as the final ${}^C_G\mathbf{T}$, and reject any hypothesis that has too few.
After this, any objects that have valid PnP poses but are not yet initialized in the scene are given an initial pose ${}^G_O\mathbf{T} = {}^C_G\mathbf{T}^{-1} {}^C_O\mathbf{T}_{pnp}$.

Since each object is initialized with a PnP pose, it is possible that the initialization can be very poor from a PnP failure, and, if the pose is bad enough (e.g., off by a large orientation error), optimization can not fix it due to only reaching local minima.
To address this issue, we check if the PnP pose from the current image yields more inliers over the last few views than the current estimated pose, and, if this is true, we re-initialize the object with the new pose.
After this, we perform a quick local refinement of the camera pose by fixing the object poses and optimizing just the current camera to better register it into the scene.

\section{Keypoint Labeling} \label{sec:supp:labeling}

\paragraph{Choice of keypoints.}
The choice of keypoints for the network to learn is important, but there is no general consensus about which choice is best.
Some have proposed to detect the corners of the 3D bounding boxes \cite{Rad2017ICCV}, while others chose keypoints that lie on the object \cite{Pavlakos2017ICRA,Peng2019CVPR} -- which seems to be the more accurate approach \cite{Peng2019CVPR}.
Inspired by \cite{Xiang2014WACV}, we try to pick keypoints that carry some semantic meaning. 
Our keypoint labels on the YCB-Video dataset can be seen in Fig. \ref{fig:ycbv_labels},
and Fig. \ref{fig:tless_labels} for the T-LESS dataset.
Specifically, we split the objects into three categories based on the overall shape -- box-like, cylinder-like, and hand tool -- and choose a unified set of keypoints for each of these shape classes based on the most identifiable features.
We found that picking a set of keypoints for each one of these classes can accurately
describe the shape of the objects for the YCB-Video and T-LESS  datasets, and the keypoint network had a relatively easy time learning the keypoints despite the fact that the shapes of some objects are not exactly rectangular, cylindrical, etc. 
In order to increase the number of keypoints and their potential usefulness in a downstream application, we also add some instance-specific keypoints, such as brand names, bar codes, and hand grips, which only show up in the YCB-Video dataset. 
Such keypoints can still be shared among multiple instances of objects in the YCB-Video dataset, but sometimes occur between shape classes (e.g., bar codes show up on the box-like cracker box and also the cylindrical soup can).

\paragraph{Labeling tool.}
To label the keypoints, we create a simple labeling program 
which allows the user to pick the same keypoint (say keypoint $k$) multiple times on the CAD model, and takes the average 3D location in the CAD model frame as the final 3D keypoint location ${}^O\bm{p}_k$.
The tool also allows the user to pick the canonical view $\{O_c\}$ used in Eq. \ref{eq:sym_gt} by simply rotating the object into the correct view.
This is especially important in the YCB-Video dataset, where the object models are not already rotated into a canonical view as they are for T-LESS.
The labeling program will be included along with our keypoint labels in the software release, which will be made available upon publication of this work.
Detailed instructions for how to reproduce our keypoint labels will also be included in this release (i.e., the rules we used to determine where each keypoint goes), which can also 
be used to label keypoints on other datasets with objects similar to YCB-Video and T-LESS.
We found that, after the user is acquainted with the labeling program, 
it only takes a few minutes per object to label the keypoints.
In the future, we would like to reduce the labeling task for the shape class-specific keypoints,
since there should be a simple set of heuristics to automatically label these when given the CAD model in a canonical
view.

\section{Extended Results} \label{sec:supp:ext}

\begin{table*} [t]
\centering
\caption{Detailed results on the YCB-Video dataset. Bold blue objects are symmetric.}
\begin{tabular}{l|cc|cc|cc|cc|cc}
\toprule
 & \multicolumn{2}{c}{PoseCNN \cite{Xiang2018RSS}} & \multicolumn{2}{c}{DeepIM \cite{Li2018ECCV}} & \multicolumn{2}{c}{PoseRBPF \cite{Deng2019RSS}} & \multicolumn{2}{c}{MHPE \cite{Fu2021IROS}}  & \multicolumn{2}{c}{Ours}  \\
\hline
Objects & \multicolumn{1}{l}{ADD} & \multicolumn{1}{l}{ADD-S} & \multicolumn{1}{l}{ADD} & \multicolumn{1}{l}{ADD-S} & \multicolumn{1}{l}{ADD} & \multicolumn{1}{l}{ADD-S} & \multicolumn{1}{l}{ADD} & \multicolumn{1}{l}{ADD-S} & \multicolumn{1}{l}{ADD} & \multicolumn{1}{l}{ADD-S}  \\
\hline                         
002\_master\_chef\_can           & 50.9 & 84.0  & 71.2 & 93.1             & 63.3 & 87.5       & 67.9 & {\bf 93.8}  & {\bf 75.0} & 87.8 \\
003\_cracker\_box                & 51.7 & 76.9  & 83.6 & {\bf 91.0}       & 77.8 & 87.6       & 67.8 & 82.9        & {\bf 84.0}  & 90.6 \\
004\_sugar\_box                  & 68.6 & 84.3  & {\bf 94.1} & {\bf 96.2} & 79.6 & 89.4       & 83.1 & 91.3        & 86.4 & 91.5 \\
005\_tomato\_soup\_can           & 66.0 & 80.9  & {\bf 86.1} & 92.4       & 73.0 & 83.6       & 79.5 & 92.2        & 85.3 & {\bf 93.5}  \\
006\_mustard\_bottle             & 79.9 & 90.2  & 91.5 & 95.1             & 84.7 & 92.0       & 81.6 & 90.8        & {\bf 94.2} & {\bf 96.2} \\
007\_tuna\_fish\_can             & 70.4 & 87.9  & {\bf 87.7} & {\bf 96.1} & 64.2 & 82.7       & 78.0 & 92.5        & 84.3  & 92.7  \\
008\_pudding\_box                & 62.9 & 79.0  & 82.7 & 90.7             & 64.5 & 77.2       & 45.4 & 71.5        & {\bf 84.1} & {\bf 92.4} \\
009\_gelatin\_box                & 75.2 & 87.1  & 91.9 & 94.3             & 83.0 & 90.8       & 76.1 & 87.8        & {\bf 94.0} & {\bf 95.9} \\
010\_potted\_meat\_can           & 59.6 & 78.5  & 76.2 & 86.4             & 51.8 & 66.9       & 69.1 & 85.5        & {\bf 83.7} & {\bf 91.7} \\
011\_banana                      & 72.3 & 85.9  & 81.2 & 91.3             & 18.4 & 66.9       & {\bf 87.7} & 93.7  & 87.3 & {\bf 94.3} \\
019\_pitcher\_base               & 52.5 & 76.8  & {\bf 90.1} & {\bf 94.6} & 63.7 & 82.1       & 76.8 & 88.8        & 89.4 & 93.9 \\
021\_bleach\_cleanser            & 50.5 & 71.9  & {\bf 81.2} & {\bf 90.3} & 60.5 & 74.2       & 47.7 & 70.3        & 61.7  & 70.5  \\
{\blue 024\_bowl}                & 6.5 & 69.7   & 8.6 & 81.4              & 28.4 & {\bf 85.6} & {\bf 40.2} & 80.1  & 32.8  & 76.9 \\
025\_mug                         & 57.7 & 78.0  & 81.4 & 91.3             & 77.9 & 89.0       & 40.6 & 72.8        & {\bf 84.8}  & {\bf 92.6} \\
035\_power\_drill                & 55.1 & 72.8  & {\bf 85.5} & {\bf 92.3} & 71.8 & 84.3       & 39.5 & 71.2        & {\bf 85.5}  & 92.2 \\
{\blue 036\_wood\_block}         & 31.8 & 65.8  & 60.0 & 81.9             & 2.3 & 31.4        & {\bf 64.6} & 85.5  & 0.0 & {\bf 86.3} \\
037\_scissors                    & 35.8 & 56.2  & 60.9 & 75.4             & 38.7 & 59.1       & 64.5 & 88.9        & {\bf 79.2} & {\bf 91.2} \\
040\_large\_marker               & 58.0 & 71.4  & 75.6 & 86.2             & 67.1 & 76.4       & 81.1 & 90.6        & {\bf 84.9} & {\bf 94.7} \\
{\blue 051\_large\_clamp}        & 25.0 & 49.9  & 48.4 & 74.3             & 38.3 & 59.3       & {\bf 49.2} & 70.7  & 47.2 & {\bf 83.0} \\
{\blue 052\_extra\_large\_clamp} & 15.8 & 47.0  & 31.0 & 73.3             & 32.3 & 44.3       & 8.6 & 47.4         & {\bf 86.3} & {\bf 94.1} \\
{\blue 061\_foam\_brick}         & 40.4 & 87.8  & 35.9 & 81.9             & 84.1 & 92.6       & 75.1 & 92.6        & {\bf 87.4}  & {\bf 93.8} \\
\hline
Mean                             & 51.7 & 75.3  & 71.7 & 88.1             & 58.4 & 76.3       & 63.1 & 82.9        & {\bf 76.1} & {\bf 90.3} \\
\bottomrule
\end{tabular}
\label{table:ycb_full}
\end{table*}

\begin{figure*}
    \centering
    \begin{subfigure}[b]{0.49\textwidth}
        \includegraphics[width=\textwidth]{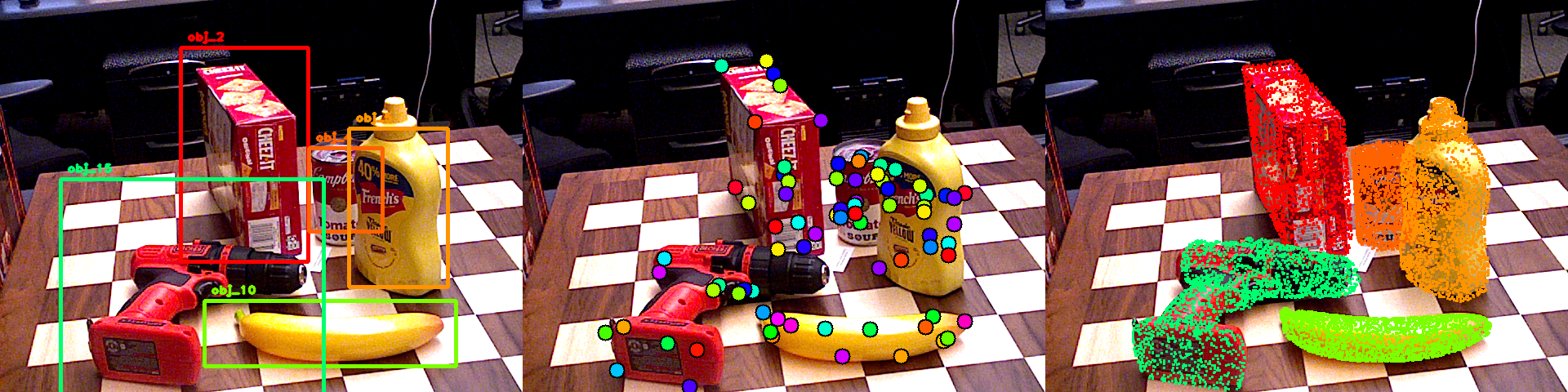}
        \includegraphics[width=\textwidth]{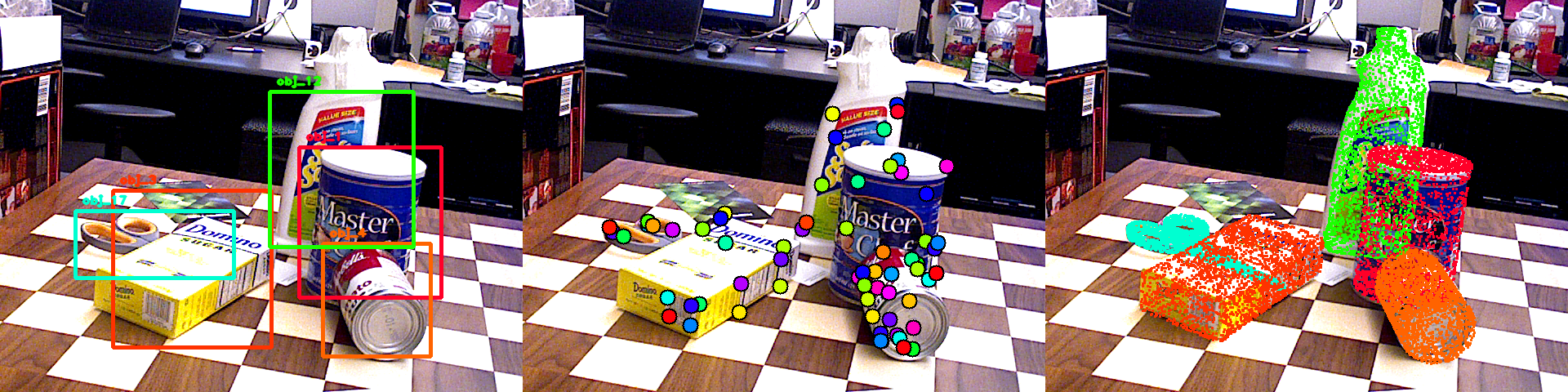}
        \includegraphics[width=\textwidth]{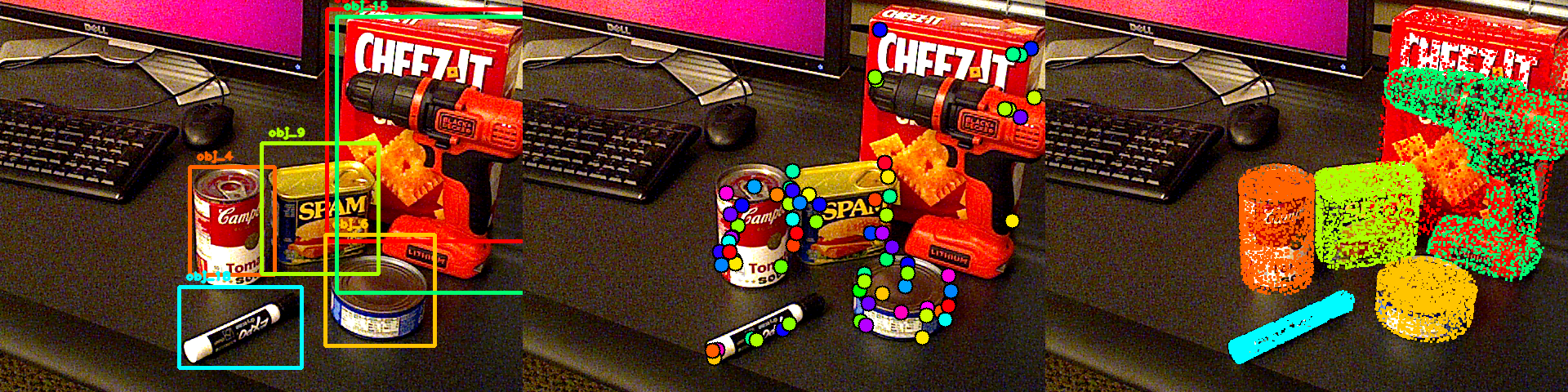}        \includegraphics[width=\textwidth]{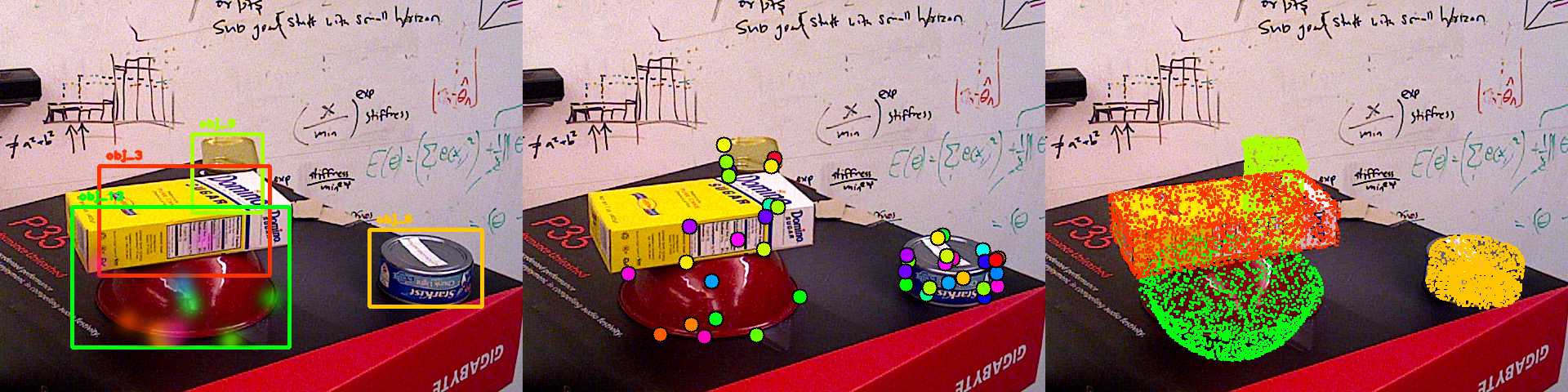}
    \end{subfigure}    
    \begin{subfigure}[b]{0.49\textwidth}
        \includegraphics[width=\textwidth]{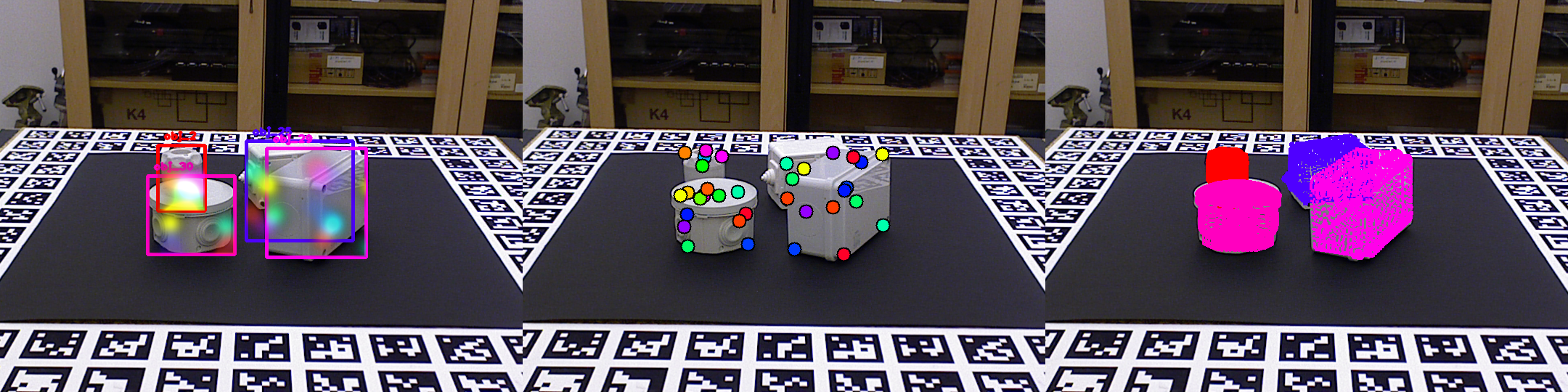}
        \includegraphics[width=\textwidth]{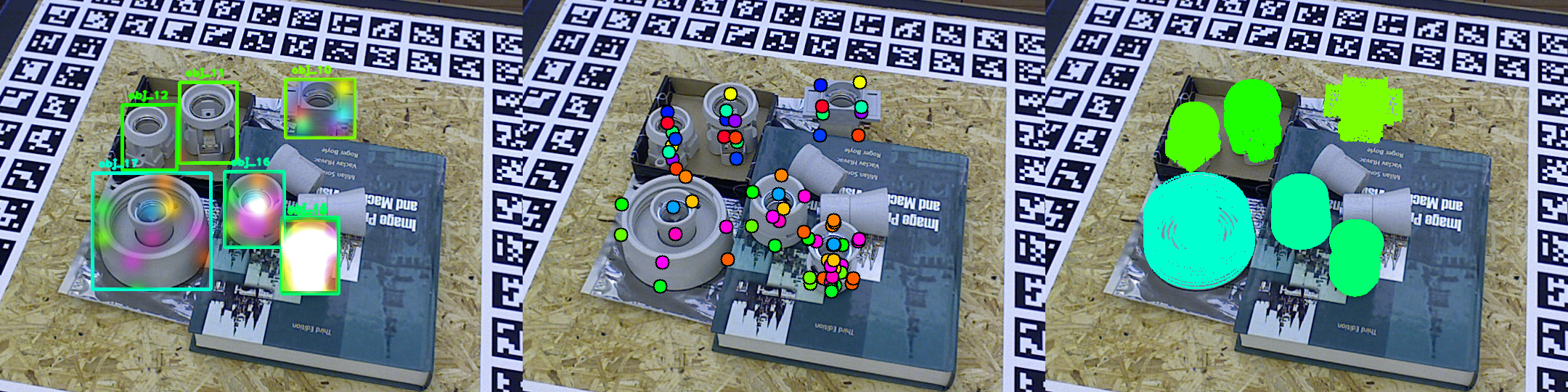}
        \includegraphics[width=\textwidth]{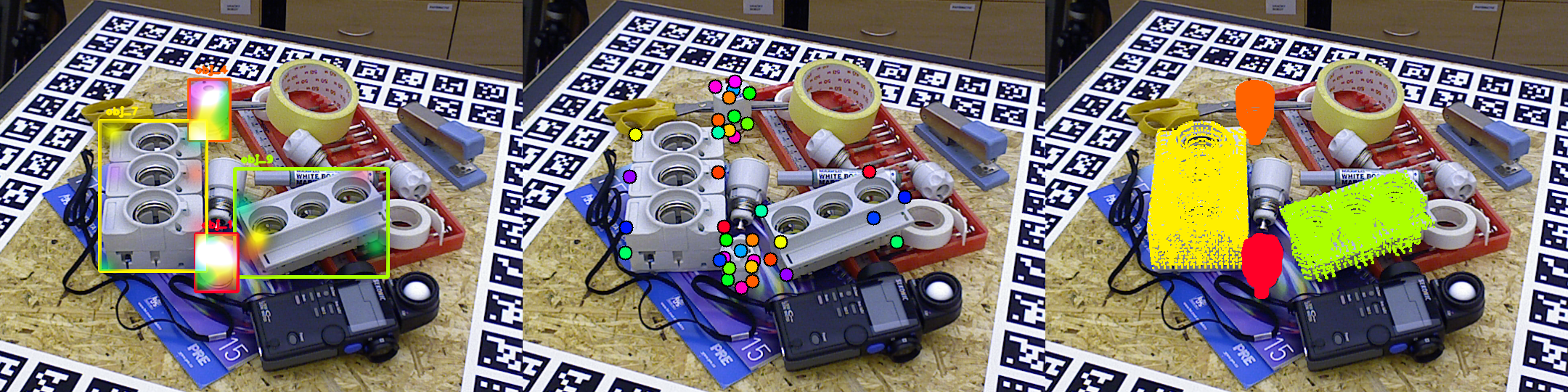}         \includegraphics[width=\textwidth]{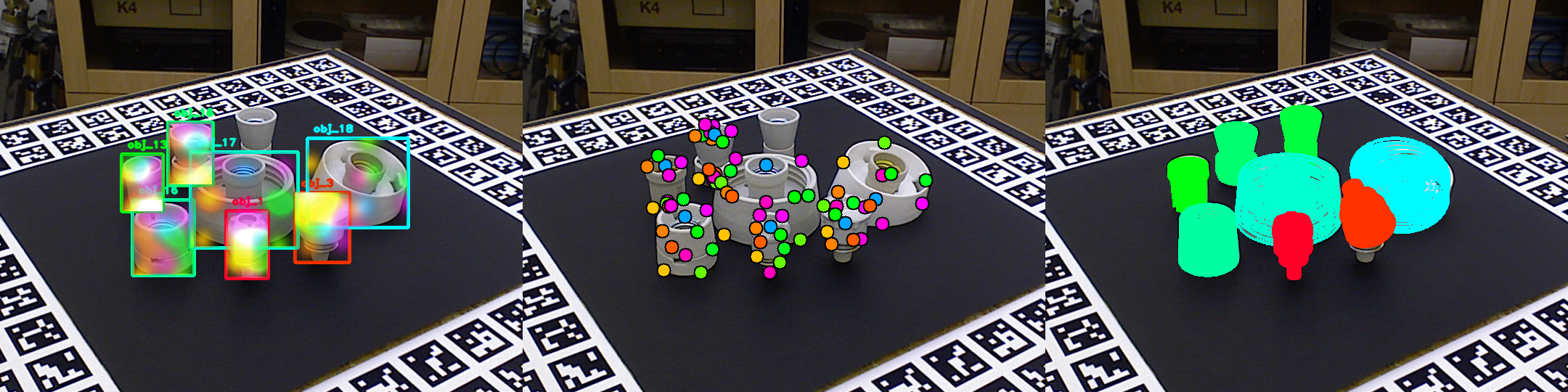}
    \end{subfigure}    
    \caption{
        Supplementary qualitative results for the YCB-Video (left) and T-LESS (right) datasets.
        The top three rows show some successful pose estimates from our system while the bottom row 
        shows a failure case. 
        The failure in both cases is from initializing objects upside down.
        The bowl on the bottom left and the orange object on the bottom right is upside down while the  are upside down.
    }
    \label{fig:qual_ext}
\end{figure*}

\paragraph{YCB-Video per-object results.}
As mentioned in Sec. \ref{sec:ycbv}, we provide more detailed results for each object on the YCB-Video dataset.
The results are presented in Table \ref{table:ycb_full}.
Here our method displays superior AUC of ADD and ADD-S for the majority of the objects.
For the five symmetric objects, which are highlighted in bold blue in Table \ref{table:ycb_full},
our method has the best AUC of ADD-S for four of them -- which shows our ability
to handle these symmetric objects effectively.
Note that the ADD metric is not very important for symmetric, since it checks for the match
to the actual ground truth pose -- which is arbitrary due to the symmetry -- while the ADD-S simply checks if the shape of the object matches well between the ground truth and estimated poses \cite{Xiang2018RSS}.
This is clear especially for the case of the wood block, where our method actually scores
a 0.0 AUC of ADD, while beating all other methods in the AUC of ADD-S metric.
This is because our estimated pose for this object correctly aligned the CAD model to the
scene to match the shape, but with a symmetry transform that yielded a completely different orientation from the ground truth.

\paragraph{Qualitative results.}
More qualitative results are shown in Fig. \ref{fig:qual_ext}.
Here we show three success cases and one failure case for both the YCB-Video and T-LESS datasets.
Our system is able to estimate correct poses for a wide variety
of difficult objects even in the presence of occlusion and bad or missing detections.
A common failure case that we saw is the system initializing objects (especially symmetric ones) upside down.
While we showed the only such case we found in the YCB-Video dataset, this is especially common in the T-LESS dataset where it is harder to distinguish the top from the bottom for many objects.
Reliably solving such edge cases is an interesting question to answer in future research. 
\end{document}